\documentclass[sigconf, screen, nonacm]{acmart}

\usepackage{mathtools}
\usepackage{amsthm}
\usepackage{verbatim}
\usepackage{multirow}
\usepackage{makecell}
\DeclareUnicodeCharacter{0301}{\'{e}}

\usepackage{colortbl}

\definecolor{navyblue}{RGB}{191, 209, 229} 
\definecolor{light_yellow}{RGB}{255,243,194}
\definecolor{orange}{RGB}{255,200,100}
\definecolor{red}{RGB}{255, 0, 0}
\definecolor{green}{RGB}{0, 176, 80}

\usepackage[capitalize,noabbrev]{cleveref}

\theoremstyle{plain}

\theoremstyle{definition}

\theoremstyle{remark}

\usepackage[textsize=tiny]{todonotes}

\usepackage{etoolbox}    
\usepackage{overpic}
\usepackage{pifont}

\definecolor{cvprblue}{rgb}{0.19,0.40,0.65}
\newcommand{\cmark}{\ding{51}} 
\newcommand{\xmark}{\ding{55}} 
\usepackage{tcolorbox}
\tcbuselibrary{breakable}
\tcbuselibrary{skins} 

\usepackage{algorithm}
\usepackage{algpseudocode}
\algrenewcommand{\algorithmicrequire}{\textbf{Input:}}
\algrenewcommand{\algorithmicensure}{\textbf{Output:}}
\makeatletter

\renewcommand\footnotetextcopyrightpermission[1]{} 
\authorsaddresses{}

\AtBeginDocument{%
  }

\begin{document}

\title{FlowScene: Style-Consistent Indoor Scene Generation with Multimodal Graph Rectified Flow}

\author{%
  Zhifei Yang$^{1*}$, Guangyao Zhai$^{2*}$, Keyang Lu$^{1}$,
  YuYang Yin$^{3}$, Chao Zhang$^{4\dagger}$,\\
  Zhen Xiao$^{1\dagger}$, Jieyi Long$^{5}$, Nassir Navab$^{2}$, Yikai Wang$^{6}$%
}
\affiliation{%
  \institution{$^1$School of Computer Science, Peking University \quad
  $^2$Technical University of Munich}
  \institution{$^3$Beijing Jiaotong University \quad
  $^4$Beijing Digital Native Digital City Research Center}
  \institution{$^5$Theta Labs, Inc. \quad
  $^6$Beijing Normal University}
  \country{\relax}
}
\renewcommand{\shortauthors}{}

\begin{abstract}
Scene generation has extensive industrial applications, demanding both high realism and precise control over geometry and appearance.
Language-driven retrieval methods compose plausible scenes from a large object database, but overlook object-level control and often fail to enforce scene-level style coherence. 
Graph-based formulations offer higher controllability over objects and inform holistic consistency by explicitly modeling relations, yet existing methods struggle to produce high-fidelity textured results, thereby limiting their practical utility.
We present FlowScene, a tri-branch scene generative model conditioned on multimodal graphs that collaboratively generates scene layouts, object shapes, and object textures. At its core lies a tight-coupled rectified flow model that exchanges object information during generation, enabling collaborative reasoning across the graph.
This enables fine-grained control of objects' shapes, textures, and relations while enforcing scene-level style coherence across structure and appearance.
Extensive experiments show that FlowScene outperforms both language-conditioned and graph-conditioned baselines in terms of generation realism, style consistency, and alignment with human preferences.
\end{abstract}

\begin{teaserfigure}
  \includegraphics[width=\textwidth]{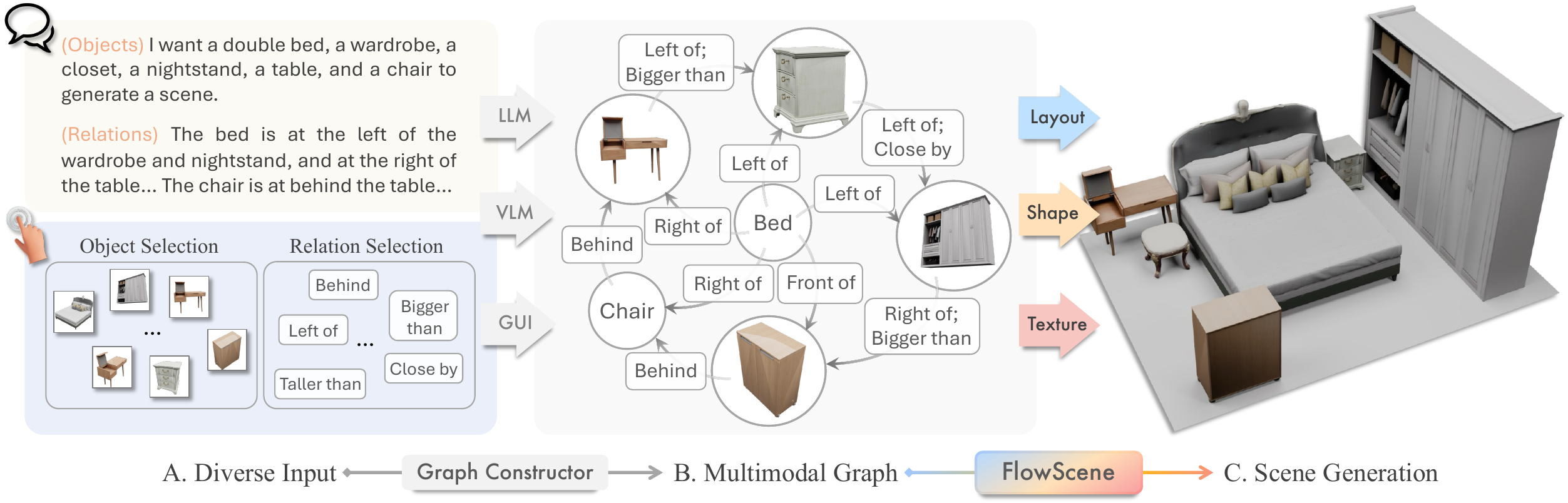}
  \caption{\textbf{Scene Generation from Diverse Input.} The prospective system, powered by \emph{FlowScene}, supports the generation of style-consistent 3D scenes from multi-source descriptions, including text input, GUI selections, and mixed information. Users can flexibly specify object categories and, if desired, provide their detailed RGB images in arbitrary views, along with their optional inter-object relationships (A). These descriptions are parsed into a multimodal graph by modern LLMs or VLMs as graph constructors (B). Based on this graph, FlowScene generates complete textured scenes, comprising both spatial scene layout and individual posed objects (C).}
  \Description{}
  \label{fig:teaser}
\end{teaserfigure}

\maketitle
\vspace{-5mm}
\noindent\rule{0.40\columnwidth}{0.4pt}
\begin{flushleft}
{\footnotesize
$^*$ Equal contribution.\\
$^\dagger$ Corresponding authors.
}
\end{flushleft}

\section{Introduction}
\label{sec:intro}

Scene generation from user prompts supports a wide spectrum of applications in manufacturing and interior design~\cite{xia2024scenegenagent}, VR/AR content creation~\cite{bautista2022gaudi}, autonomous driving~\cite{pronovost2024scenario}, and robotics~\cite{jiang2024roboexp}. These settings demand high realism and precise control over geometry and appearance, allowing users to specify object categories, semantic and spatial relations, and desired individual appearances. The scene generator must follow these instructions while preserving scene-level style consistency across object structure and visual appearance. A prospective workflow is able to transfer flexible and diverse inputs into a high-quality scene (cf.~\autoref{fig:teaser} \textcolor{cvprblue}{A} and \textcolor{cvprblue}{C}).

Training-free approaches~\cite{yang2024holodeck,sun2025layoutvlm} that rely directly on large models fail to meet this goal. They typically retrieve well-designed meshes to compose scenes from coarse language commands, yet provide limited per-object control, overlook inter-object relations, and rarely maintain scene-level style consistency. This leads to scale, topology, and appearance mismatches in the final scenes. For more interactive workflows, scene-graph–based methods~\cite{planner3d,zhai2024commonscenes,zhai2024echoscene,yang2025mmgdreamer} explicitly model objects and their relations, enabling strong per-object controllability, better handling of relational constraints, and improved structural consistency. 
However, these methods fail to generate textured scenes in an end-to-end manner, which results in low-fidelity generation and limits their utility in downstream tasks.

Simultaneously achieving high fidelity and maintaining style consistency across objects under flexible control is non-trivial, especially when some objects lack visual or textual cues. In such cases, the model must aggregate scene-level context to guide the generation of their geometry and appearance.
In this paper, we introduce \emph{FlowScene} for compositional scene generation that maintains high fidelity and style consistency. 
As shown in~\autoref{fig:teaser} \textcolor{cvprblue}{B-C}, FlowScene accepts a multimodal scene graph in which each node can fuse textual and visual information of the object, and employs three branches to generate scene layouts, object shapes, and object textures, respectively.  
At the core of each branch is \emph{Multimodal Graph Rectified Flow}, which tightly couples exchanges of node information. Specifically, every node carries denoising states matched to each branch, including bounding boxes for the layout branch, voxelized latents for the shape branch, and structured latents for the texture branch. Throughout the denoising steps, nodes interact iteratively with each other along graph edges, ensuring progressively refined results and yielding fine-grained control over per-object appearance, inter-object spatial relations, and cross-object style consistency across structure and texture. Textured shapes are finally populated into the generated layout to produce the full scene.

We train FlowScene on 3D-FRONT~\cite{fu20213d} with SG-FRONT~\cite{zhai2024commonscenes}. Experimental performance shows that our results outperform language-driven synthesis~\cite{yang2024holodeck,sun2025layoutvlm} and graph-conditioned generation~\cite{yang2025mmgdreamer} in realism, style consistency, and human preference. Moreover, FlowScene significantly accelerates the graph-conditioned generation process while enhancing both the quality of individual objects and the overall performance of holistic scene generation. 
For the long-horizon application scope, we apply FlowScene as a robust backend that initiates scene creation based on user-provided sentences, or interactive selections of objects and relationships, or both. \textbf{Our contributions are summarized as follows:} 
\begin{itemize}
    \item We present \emph{FlowScene} for generating high-fidelity 3D scenes from multimodal graphs, supporting fine-grained control in object-level appearance and scene-level style consistency across structure and texture.
    \item We detail the core of FlowScene as \emph{Multimodal Graph Rectified Flow}. It exchanges node information during sampling to satisfy both individual and holistic conditions, and achieves faster generation than the previous diffusion-based mechanism.
    \item We show stronger performance than competitors in generation realism, style consistency, and human preference alignment. We  provide a workflow of how FlowScene facilitates scene generation from diverse input sources.
\end{itemize}

\section{Related Work}
\label{sec:rw}

\paragraph{Scene Graph and Applications}
Scene graphs provide a symbolic representation of a scene as a graph with object nodes and directed edges encoding inter-object relations. They can be constructed from multiple modalities, including text~\cite{zhao2023textpsg}, 2D images~\cite{qi2019attentive}, and 3D geometry~\cite{koch2024sgrec3d}, and even 4D spatio-temporal data~\cite{yang20234d}, enabling rich spatial and temporal understanding. Early works established scene graphs for retrieval and reasoning~\cite{krishna2017visual}, and subsequent research has leveraged them across a wide range of downstream tasks, including visual retrieval~\cite{johnson2015image,fang2023alignment,fang2023mask,wang2025mari}, question answering~\cite{teney2017graph}, controllable image synthesis~\cite{wu2023scene}, video synthesis~\cite{cong2023ssgvs,yang2025vidtext}, 3D scene understanding~\cite{looper20223d}, and 3D scene synthesis~\cite{yang2025mmgdreamer}. 
Scene graphs also support embodied applications such as manipulation planning~\cite{zhai2024sg} and instruction-conditioned navigation~\cite{rana2023sayplan,ma2025ctsg}. In this work, we adopt a multimodal graph formulation that integrates textual and visual information at the node level~\cite{yang2025mmgdreamer}.

\paragraph{Rectified Flow and Applications}
Rectified flow and flow matching~\cite{liu2022flow} have emerged as a strong alternative to diffusion-based generators~\cite{ho2020denoising}, leveraging straight-line supervision with deterministic ODE sampling to reduce training variance and enable few-step generation. At scale, rectified flow transformers achieve the best image quality with competitive speed and model scaling behavior~\cite{esser2024scaling}, and multi-rate designs extend these benefits to video by improving temporal coherence and long-horizon efficiency~\cite{jin2024pyramidal}. Efficient ODE solvers further amplify these gains~\cite{lu2022dpm}, while related fast-sampling paradigms provide complementary perspectives on one/few-step generation~\cite{song2023consistency}. Building on these advances, we adopt rectified flow as our backbone for 3D scene generation when coupled with the graph representation.

\paragraph{3D Scene Synthesis} 3D scene synthesis supports robotics and AR~\cite{mandlekar2023mimicgen,tahara2020retargetable} and is commonly driven by {text}~\cite{tang2023diffuscene,pun2025hsm,fu2024anyhome,ma2024fastscene, lu2025yo}, {graphs}~\cite{zhai2024commonscenes,zhai2024echoscene,gao2024graphdreamer}, or {images}~\cite{huang2024midi,ling2025scenethesis}. Text-conditioned methods leverage LLM priors to generate layouts or full scenes~\cite{yang2024holodeck,li2024dreamscene,song2023roomdreamer,ocal2024sceneteller}, but often yield incomplete or ambiguous spatial structure without strong visual grounding~\cite{feng2024layoutgpt,yang2024llplace,ccelen2025design,aguina2024open}. Graph-conditioned pipelines encode objects and relations to improve coherence and control~\cite{dhamo2021graph,lin2024instructscene}, including relation conditioning~\cite{zhai2024commonscenes,zhai2024echoscene}, with hierarchical extensions~\cite{sun2025hierarchically}. Image-based approaches exploit visual priors~\cite{wang2025architect,wang2023perf,hara2024magritte,hollein2023text2room}, but fixed viewpoints limit holistic 3D reasoning and introduce cross-view inconsistencies~\cite{sun2025layoutvlm,deng2025global}. Along the modeling axis, works span autoregressive generators~\cite{wang2021sceneformer,paschalidou2021atiss,zhao2024roomdesigner}, layout/shape priors~\cite{engelmann2021points,xu2023discoscene,yan2024frankenstein,jyothi2019layoutvae,zhang2024style,epstein2024disentangled}, and diffusion-style objectives~\cite{yang2024physcene,meng2024lt3sd,hollein2023text2room,wu2024blockfusion}. Despite progress, current systems still do not unify multimodal inputs while offering both geometric and appearance control and reliable relation compliance, which is the goal of FlowScene in this paper.

\section{Preliminary}
\label{sec:pre}

\paragraph{Rectified Flow}
Rectified flow~\cite{liu2022flow} is the straight-line instantiation of flow matching models~\cite{lipman2022flow}, which learns a time-dependent velocity field \(v_\theta(d,t)\), \(t\in[0,1]\), such that the ODE \(\dot{d}_t = v_\theta(d_t,t)\) transports the data distribution \(p_{\mathrm{data}}\) to a simple prior \(p_1\)(e.g., \(\mathcal N(0,I)\)), where \(d\in\mathbb{R}^n\) denotes the data variable. Training uses linear paths between \(d_0\) from the data and \(d_1\) from the prior, with identity path parameterization \(d_t=(1-t)d_0+t d_1\). The target velocity is constant along each path to be \(d_1-d_0\).
We optimize the model by least-squares regression to these targets,
\begin{equation}
    \mathcal L_{\mathrm{RF}}(\theta)=\mathbb{E}_{\,d_0,\,d_1,\,t}\!\left[\big\|v_\theta(d_t,t)-(d_1-d_0)\big\|_2^2\right],
\end{equation}
where \(d_0\!\sim\!p_{\mathrm{data}}\), \(d_1\!\sim\!p_1\), and \(t\) is sampled from a LogNormal(1,1) derived schedule on \([0,1]\). During sampling, we draw \(d_1\!\sim\!p_1\) and integrate the reverse-time ODE \(\dot{d}_t = -v_\theta(d_t,t)\)  from \(t{=}1\) to \(0\). Optional conditioning on side information \(c\) (e.g., text, camera, geometry) is handled by using \(v_\theta(d,t\mid c)\) and forming the same straight-line targets with \(c\) included; the objective keeps the same form.

\paragraph{Multimodal Graph}
A scene graph \(\mathcal G=(\mathcal V,\mathcal E)\) represents an environment with object nodes and their relationships~\cite{chang2021comprehensive}. The node set \(\mathcal V=\{v_i\mid i=1,\ldots,N\}\) carries categorical embeddings \(p_i\), and the edge set \(\mathcal E=\{e_{i\to j}\mid i,j\in\{1,\ldots,N\},\, i\neq j\}\), where \(|\mathcal E|=M\), carries predicate labels \(\tau_{i\to j}\), indicating relations from \(v_i\) to \(v_j\).
To capture richer cues than semantics alone, a multimodal scene graph is introduced by~\cite{yang2025mmgdreamer} as \(\mathcal{G}_\mathrm{M}=(\mathcal V_\mathrm{M},\mathcal E)\), and its $i$-th node \(v_i^m = [p_i,u_i,f_i]\) aggregates a learnable embedding \(p_i\) with foundation text features \(u_i\) (e.g., CLIP~\cite{radford2021learning}), or foundation visual features \(f_i\) (e.g., CLIP/DINOv2~\cite{oquab2023dinov2} from the object image) or both. Missing modalities are zero-padded to a common format (e.g., \([p_i,u_i,f_i]\mapsto[p_i,u_i,\mathbf 0]\) or \([\mathbf 0,\mathbf 0,f_i]\)). \(\mathcal{G}_\mathrm{M}\) thereby supports nodes that are (i) text-only, (ii) image-only, or (iii) multimodal, allowing the graph to unify language-only inputs (parsed by an LLM), GUI/VLM-derived visual nodes, or their mixtures, as shown in~\autoref{fig:teaser} \textcolor{cvprblue}{A-B}. 

\(\mathcal{G}_\mathrm{M}\) can be encoded by a \(L\)-layer triplet Graph Convolutional Network (triplet-GCN)~\cite{johnson2018image} \(\Omega_{\mathcal G}\) for message passing and aggregation. Let the node features after the \(l\)-th layer be \(\delta^{(l)}_i\) with \(\delta^{(0)}_i=v_i^m\). Each layer updates via
\begin{subequations}\label{eq:triplet-gcn}
\begin{gather}
(\gamma^{(l)}_{i}, \tau^{(l+1)}_{i \to j}, \gamma^{(l)}_{j})
= \operatorname{MLP}\!\big(\delta^{(l)}_{i},\, \tau^{(l)}_{i \to j},\, \delta^{(l)}_{j}\big),
\label{eq:mp}\\[2pt]
\delta^{(l+1)}_{i}
= \gamma^{(l)}_{i} + \operatorname{MLP}\!\left(\operatorname{Avg}\!\big(\gamma^{(l)}_{j} \mid j \in \mathcal{N}_{\mathcal G}(i)\right),
\label{eq:fa}
\end{gather}
\end{subequations}
where \(\mathcal N_{\mathcal G}(i)\) are neighbors of \(v_i^m\) and \(\operatorname{Avg}\) is mean pooling. \eqref{eq:mp} facilitates message passing between connected nodes, while \eqref{eq:fa} performs feature aggregation for each node. 

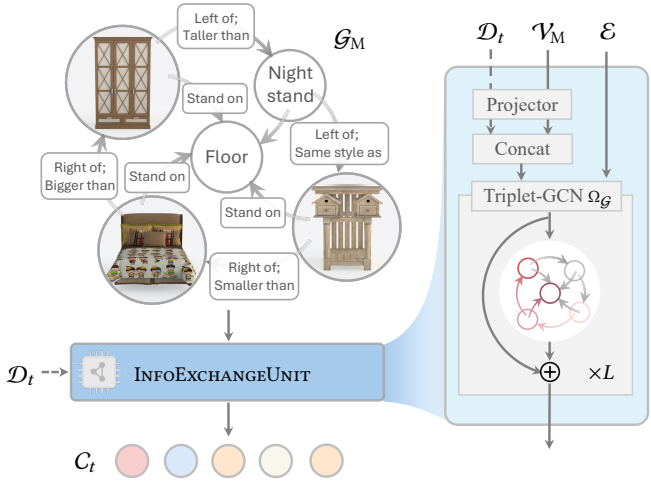
\begin{figure}[t]
  \centering
  \begin{overpic}[width=0.97\linewidth,trim=0cm 0 0 0cm,clip]{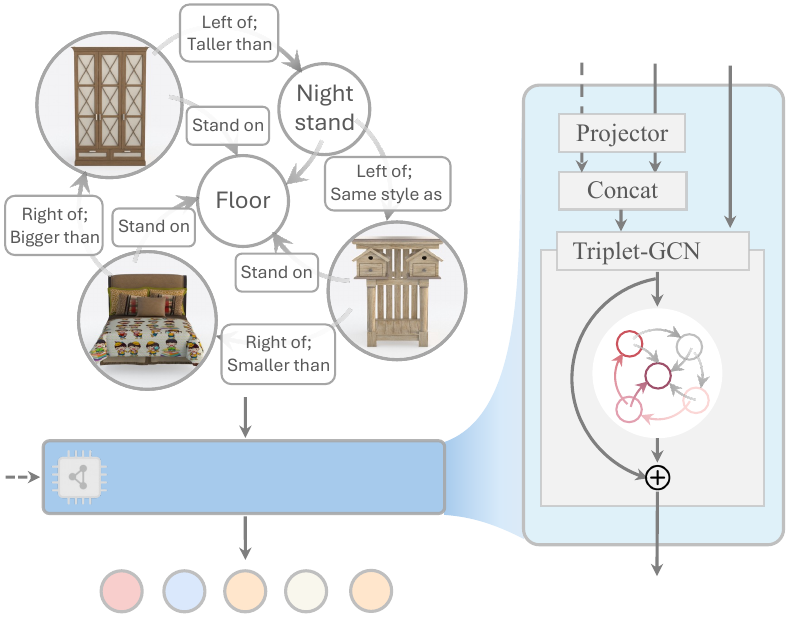}
    \put(47,71){\color{black}{ $\mathcal{G}_\mathrm{M}$}}
    \put(15.7,16.7){\color{black}{\small \textsc{InfoExchangeUnit}}}
    \put(-5,16.5){\color{black}{$\mathcal{D}_t$}}
    
    \put(71,72){\color{black}{$\mathcal{D}_t$}}
    \put(80,72){\color{black}{$\mathcal{V}_\mathrm{M}$}}
    \put(91,72){\color{black}{$\mathcal{E}$}}

    \put(89,45.6){\color{black}{\scriptsize $\Omega_\mathcal{G}$}}
    \put(89,17){\color{black}{\small $\times L$}}

    \put(6,2.5){\color{black}{$\mathcal{C}_t$}}
  \end{overpic}
    \caption{\textbf{Multimodal Graph Processing.} Given a graph $\mathcal{G}_\mathrm{M}$ (upper left), the graph network (right) first projects multimodal features of nodes $\mathcal{V}_\mathrm{M}$, followed by message passing and feature aggregation along the edges $\mathcal{E}$ connecting $\mathcal{V}_\mathrm{M}$. Within the denoising process of a generative model, it acts as an \textsc{InfoExchangeUnit} to incorporate temporal denoising data $\mathcal{D}_t$, which is projected and fused with the node features (lower left). In this way, the network yields node-wise conditioning $\mathcal{C}_t$ while respecting the global constraints imposed by the graph structure (bottom).
}
    \label{fig:gcn}
\end{figure}

\section{Method}
\label{sec:method}
We first illustrate the proposed graph rectified flow as a common graph-conditioned generation backbone. Then we delve into the details of how we integrate it into FlowScene.

\subsection{Multimodal Graph Rectified Flow}
\label{GRF}
Our formulation is essentially a tightly coupled rectified flow designed for multiple content generation. Original rectified flow~\cite{liu2022flow} is applied to single-content generation, while our variant operates on all contents jointly to achieve high single-content quality and inter-content consistency.

\paragraph{Training}
The overall procedure is summarized in~\autoref{alg:grf_train}.
We define target data samples of $N$ nodes as $\mathcal{D}_0=\{d_0^i~|~i \in \{1,  \ldots, N \}\}$ and a known prior as $\mathcal{D}_1$.
In the forward process, we deploy an $N$-thread linear interpolation $\mathcal{D}_t = (1-t)\mathcal{D}_0 + t \mathcal{D}_1$, which plays the role of the “reference” trajectories, analogous to the straight-line path connecting the source node data and the global priors, typically being white noise.
The backward process is represented as a constant vector field $v(\mathcal{D}_t, \mathcal{C}_t, t)=\mathcal{D}_1 - \mathcal{D}_0$. Here, $\mathcal{C}_t=\{c_t^i~|~i \in \{1, \ldots, N\}\}$ are the time-dependent, tightly coupled conditions, which exchange inter-object information to guide the denoising process.
To obtain $\mathcal{C}_t$, as shown in~\autoref{fig:gcn}, we adapt the triplet-GCN~\eqref{eq:triplet-gcn} to an \textsc{InfoExchangeUnit}  fed on $\mathcal{G}_\mathrm{M}$ and $\mathcal{D}_t$ to perform feature aggregation for timestep-wise condition: 
Each node now stacks the feature $v_i^m$ and the noisy $d_t^i$ as $\delta_i^d = v_i^m \oplus  \operatorname{Projector}(d_t^i),$ and thus \(\mathcal{G}_\mathrm{M}\) is augmented to $ \mathcal{G}_\mathrm{M}^\mathcal{D}$, which is processed by \eqref{eq:triplet-gcn} to reflect the global data constraints inside the conditions $\mathcal{C}_t=\Omega_\mathcal{G}(\mathcal{G}_\mathrm{M}^\mathcal{D})$. Therefore, a denoiser  $\Theta_\mathcal{D}$ is trained to approximate the vector field $v$ by minimizing the objective:
\begin{equation}
\label{loss}
    \mathcal L_{\mathrm{GRF}}= \mathbb{E}_{\mathcal{D},\mathcal{C},t}\left[\| \Theta_\mathcal{D}(\mathcal{D}_t,\mathcal{C}_t,t) - v \|_2^2\right].
\end{equation}
The architecture of $\Theta_\mathcal{D}$ can be various, and we adopt the rectified flow transformers~\cite{xiang2024structured}.

\begingroup
\setlength{\textfloatsep}{6pt} \begin{algorithm}[t]
\caption{Multimodal Graph Rectified Flow--Training}
\label{alg:grf_train}
\begin{algorithmic}[1]
\Require Graph $\mathcal{G}_\mathrm{M}$, $\mathcal{D}_0=\{d_0^i\}_{i=1}^N$, noise sampler for $\mathcal{D}_1$
\While{not converged}
    \State Sample batch $(\mathcal{G}_\mathrm{M},\mathcal{D}_0)$
    \State Calculate $\mathcal{D}_1$ based on $\mathcal{D}_0$ \Comment{$\mathcal{D}_1$ subject to $\mathcal{N}(0,1)$}
    \State Sample $t \sim \mathrm{LogNormal}(1,1)$
    \State $\mathcal{D}_t \gets (1-t)\mathcal{D}_0 + t\mathcal{D}_1$
    \State $v_{\text{target}} \gets \mathcal{D}_1 - \mathcal{D}_0$ \Comment{rectified target velocity}
    \State $\mathcal{C}_t \gets \textsc{InfoExchangeUnit}(\mathcal{G}_\mathrm{M},\mathcal{D}_t, t)$ 
    \State $\hat v \gets \Theta_\mathcal{D}(\mathcal{D}_t,\mathcal{C}_t,t)$
    \State $\mathcal L_{\mathrm{GRF}} \gets \|\hat v - v_{\text{target}}\|_2^2$ \Comment{Eq.~(\ref{loss})}
    \State Update $\Theta_\mathcal{D},\textsc{InfoExchangeUnit}$ via backprop
\EndWhile
\end{algorithmic}
\end{algorithm}
\endgroup

\begingroup
\setlength{\textfloatsep}{6pt} 
\begin{algorithm}[t]
\caption{Multimodal Graph Rectified Flow--Inference}
\label{alg:grf_infer}
\begin{algorithmic}[1]
\Require Graph $\mathcal{G}_\mathrm{M}$, sample steps $K$
\State Initialize $\mathcal{D}_1$ from noise $\mathcal{N}(0,1)$
\State $\Delta t \gets 1/K$
\For{$k = K-1$ \textbf{down to} $0$} \Comment{from $t=1$ to $t=0$}
    \State $t \gets (k+1)\Delta t$
    \State $\mathcal{C}_t \gets \textsc{InfoExchangeUnit}(\mathcal{G}_\mathrm{M},\mathcal{D}_t, t)$
    \State $\mathcal{D}_{t-\Delta t} \gets \mathcal{D}_t - \Theta_\mathcal{D}\!\left(\mathcal{D}_t,\mathcal{C}_t,\,t\right) \cdot \Delta t$
\EndFor
\State \Return $\mathcal{D}_0$
\end{algorithmic}
\end{algorithm}
\endgroup

\paragraph{Inference}
At inference time, the process starts from $\mathcal{D}_1 \sim \mathcal{N}(0,1)$. The model then integrates the learned conditional vector field, in order to evolve $\mathcal{D}_t$ from $\mathcal{D}_1$ back toward the target data distribution $\mathcal{D}_0$. Formally, the trajectory follows ${\Delta\mathcal{D}_t} = -\Theta_\mathcal{D}\bigl(\mathcal{D}_t, \mathcal{C}_t, t\bigr){\Delta t}$,
integrated from $t=1$ to $t=0$ along $K$ steps. Owing to the near-linear paths induced by rectified flow training, this integration requires only a small number of steps in practice; see~\autoref{alg:grf_infer}.


\begin{figure*}[h]
  \centering
  \begin{overpic}[width=0.95\textwidth,trim=0cm 0 0 0cm,clip]{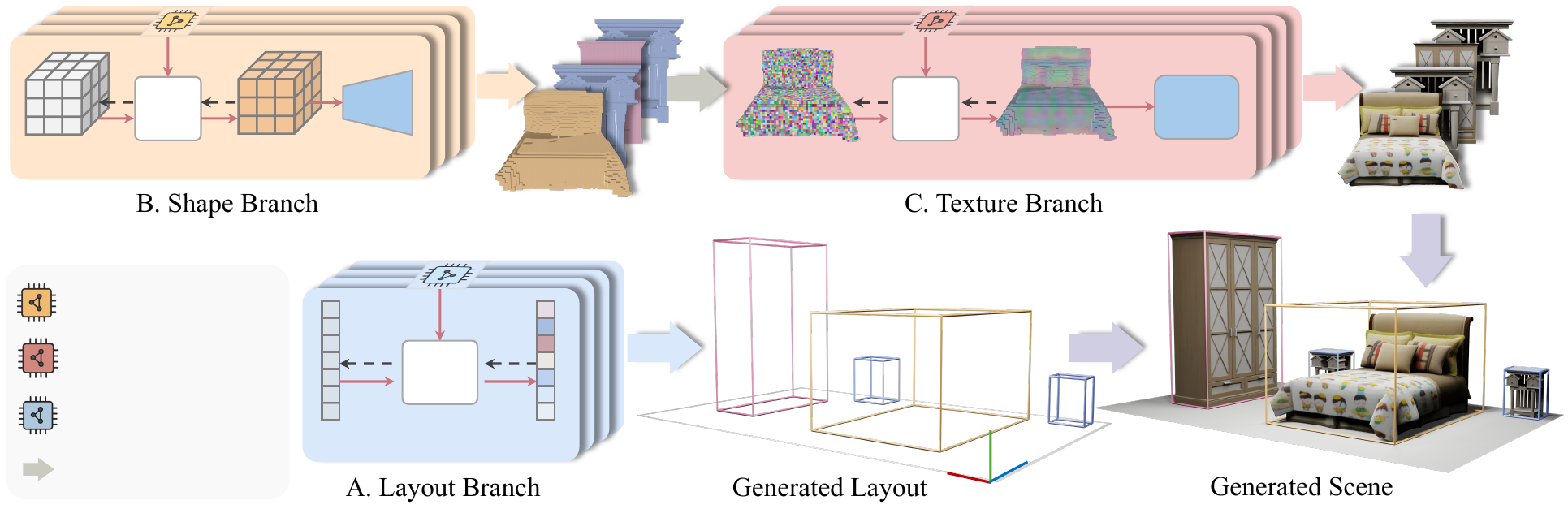}
    \put(3,22){\color{black}{\small $\mathcal{S}_1$}}
    \put(16.5,22){\color{black}{\small $\mathcal{S}$}}
    \put(9.5,25){\color{black}{\small $\Theta_\mathcal{S}$}}
    \put(22.9,25.4){\color{black}{\small $\Phi_{D}$}}
    \put(41.5,21){\color{black}{\small $\mathcal{X}$}}

    \put(49.6,21.8){\color{black}{\small $\mathcal{X}_1^e$}}
    \put(66.2,21.8){\color{black}{\small $\mathcal{X}^e$}}
    \put(57.7,25){\color{black}{\small $\Theta_\mathcal{T}$}}
    \put(75,25){\color{black}{\small $\Psi_{D}$}}
    \put(94.5,21){\color{black}{\small $\mathcal{O}$}}

    \put(20.4,3.6){\color{black}{\small $\mathcal{B}_1$}}
    \put(34.2,3.6){\color{black}{\small $\mathcal{B}$}}
    \put(27,8){\color{black}{\small $\Theta_\mathcal{B}$}}

    \put(4,12.6){\scalebox{0.85}{\scriptsize \textsc{ShapeExchangeUnit}}}
    \put(4,9){\scalebox{0.85}{\scriptsize \textsc{TextureExchangeUnit}}}
    \put(4,5.5){\scalebox{0.85}{\scriptsize \textsc{LayoutExchangeUnit}}}
    \put(4,2){\scalebox{0.85}{\small Noise Anchoring}}
    
  \end{overpic}
    \caption{
    \textbf{FlowScene Pipeline.}  Conditioned on $\mathcal{G}_\mathrm{M}$ in~\autoref{fig:gcn}, FlowScene stacks three coordinated branches to generate a scene.  \emph{A. Layout Branch} generates the scene layout from noise $\mathcal{B}_1$, where the x-axis (red) points to the right and the z-axis (blue) points to the front. The \textsc{LayoutExchangeUnit} iteratively applies temporal layout constraints to the generation process by conditioning $\Theta_\mathcal{B}$.  \emph{B. Shape Branch} is parallel to the layout branch. It exchanges shape information among nodes according to $\mathcal{G}_\mathrm{M}$ through the \textsc{ShapeExchangeUnit}. At inference, we sample noise $\mathcal{S}_1$, and generate latent shape codes $\mathcal{S}$ via $\Theta_\mathcal{S}$, and decode them into voxelized shapes $\mathcal{X}$ using $\Phi_D$.  \emph{C. Texture Branch} is subsequent to the shape branch. It anchors Gaussian noise onto $\mathcal{X}$ to form $\mathcal{X}_1^e$, which is then denoised by $\Theta_\mathcal{T}$ into structured latents $\mathcal{X}^e$. Finally, $\Psi_D$ decodes them into objects $\mathcal{O}$, which are scaled and placed into the generated layout $\mathcal{B}$ to complete the scene. 
}
    \label{fig:pipeline}
\end{figure*}
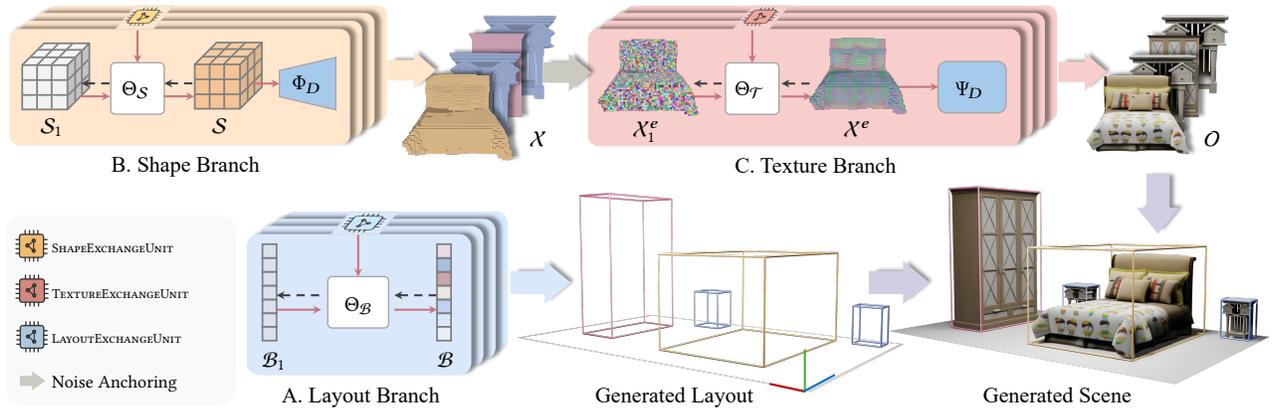

\subsection{Tri-branch Generation}
Building on the proposed graph flow model, we develop FlowScene for scene generation, as illustrated in~\autoref{fig:pipeline}. The framework consists of three branches responsible for generating the scene layout, object shapes, and object textures. The final scene is synthesized by populating the generated layouts with textured object shapes.  

\paragraph{Layout Branch}
We represent the scene layout using 3D object bounding boxes, as drawn in~\autoref{fig:pipeline}.  
Each bounding box \( \mathbf{b}_i \) is defined by its normalized location \( \mathbf{t}_i \in \mathbb{R}^3 \), size \( \mathbf{s}_i \in \mathbb{R}^3 \), and rotation angle \( \alpha_i \) expressed in sine–cosine form \( [\cos(\alpha_i), \sin(\alpha_i)] \). Thus, the complete layout is represented as  
\(
\mathcal{B} = \{[\mathbf{t}_i, \mathbf{s}_i, \cos(\alpha_i), \sin(\alpha_i)] ~|~ i \in \{1,\ldots,N\}\}.
\)
This branch is trained independently to generate $\mathcal{B}$ by optimizing Eq.~\eqref{loss}, with the denoising data following $\mathcal{D}_t \mapsto \mathcal{B}_t$ and thus the data graph transforming into a layout-enhanced graph $\mathcal{G}_\mathrm{M}^\mathcal{D} \mapsto \mathcal{G}_\mathrm{M}^\mathcal{B}$. In the forward process, Gaussian noise is iteratively added until $\mathcal{B}$ becomes white noise as $\mathcal{B}_1$. The denoiser $\Theta_\mathcal{D} \mapsto \Theta_\mathcal{B}$ is then conditioned on a \textsc{LayoutExchangeUnit}, specialized from the \textsc{InfoExchangeUnit} which extracts global layout constraints from $\mathcal{G}_\mathrm{M}^\mathcal{B}$.

\paragraph{Shape Branch}
Following previous work~\cite{zhai2024commonscenes,xiang2024structured}, training data for this branch are prepared by voxelizing objects into a sparse structure $\mathcal{X}=\{x_i~|~ i \in \{1,\ldots,N\}\}$. We use a shape VQ-VAE $\Phi_{(E,D)}$~\cite{van2017neural}, which encodes $\mathcal{X}$ into compact latent code $\mathcal{S}=\{s_i~|~ i \in \{1,\ldots,N\}\}$. We show the procedure in~\autoref{sec:object},~\autoref{fig:process} \textcolor{cvprblue}{A–B}. Since $|s_i| \ll |x_i|$, $\Phi$ provides computationally efficient modeling.  
Similar to the layout branch, this branch is trained independently with Eq.~\eqref{loss}, with noise iteratively added in the forward process and mappings $\mathcal{D}_t \mapsto \mathcal{S}_t$, $\mathcal{G}_\mathrm{M}^\mathcal{D} \mapsto \mathcal{G}_\mathrm{M}^\mathcal{S}$, and $\Theta_\mathcal{D} \mapsto \Theta_\mathcal{S}$.
During denoising, the \textsc{InfoExchangeUnit} is specialized into a \textsc{ShapeExchangeUnit}, which exchanges shape information according to $\mathcal{G}_\mathrm{M}^\mathcal{S}$ to ensure consistent generation compliant with high-level edges (e.g., \texttt{the same style as}).  
The branch ultimately generates $\mathcal{S}$ via $\Theta_\mathcal{S}$, and decodes them into shapes $\mathcal{X}=\Phi_D(\mathcal{S})$.  
 
\paragraph{Texture Branch}
The texture branch is subordinate to the shape branch, as an object’s texture is anchored to its geometry. Nevertheless, it is trained independently using the same rectified flow objective in Eq.~\eqref{loss}, with branch-specific denoiser and exchange unit. Following Trellis~\cite{xiang2024structured}, we render each object from random spherical views, and extract features with a pre-trained DINOv2 encoder~\cite{oquab2023dinov2}. Each voxel in $x_i$ projects onto the multiview features, which are averaged into $x_i^f$. A texture VQ-VAE $\Psi_{(E,D)}$ learns to reconstruct the object, modeling a structured latent $x_i^e$ (cf.~\autoref{sec:object},~\autoref{fig:process} \textcolor{cvprblue}{A–C}).  
Thus, $\mathcal{D}_0$ maps to $\mathcal{X}^e = \{x_i^e ~|~ i \in \{1,\ldots,N\}\}$. Unlike the other branches, here noise is iteratively added only to the features of $\mathcal{X}^e$, while the geometric structure remains unchanged. Note that $\mathcal{X}^e$ shares the same geometry as $\mathcal{X}$. Mappings follow $\mathcal{G}_\mathrm{M}^\mathcal{D} \mapsto \mathcal{G}_\mathrm{M}^\mathcal{T}$ and $\Theta_\mathcal{D} \mapsto \Theta_\mathcal{T}$ with \textsc{InfoExchangeUnit} specialized into a \textsc{TextureExchangeUnit}. 
At inference, the branch takes decoded shapes $\mathcal{X}$ and initializes $\mathcal{X}_1^e$ by anchoring Gaussian noise to the structured latent while keeping the geometric structure fixed. 
During the denoising process, the \textsc{TextureExchangeUnit} exchanges texture information among nodes to form $\mathcal{X}^e$, resulting in style-consistent generation $\mathcal{O}$. 
This procedure is particularly important for text-only nodes, whose appearances are primarily inferred from the exchanged information.

\section{Experiments}
\label{sec:exp}
We first describe the experimental setup, including the dataset, baselines within different technical routes, metrics with a perceptual study for each aspect of generation quality, and implementation details. We then present quantitative and qualitative results against these baselines, followed by an ablation study on the architecture and input to assess their effect of generation fidelity and style consistency.

\begin{table*}[t!]
\caption{\textbf{Scene-Level Fidelity.} This is evaluated by comparing top-down renderings of generated and real scenes at $256^{2}$ resolution using FID, $\text{FID}_{\text{CLIP}}$, and KID ($\times 0.001$), following~\cite{zhai2024echoscene} (lower is better). * denotes their retrieval branches, which generate layouts and retrieve well-designed meshes from 3D-FUTURE~\cite{fu20213d_future}. The best results are in \textbf{bold}.}
\centering
    \scalebox{1}{
    \begin{tabular}{l  c c c c c c c c c }
     \toprule 
        \multirow{2}{*}{Method} & \multicolumn{3}{c}{Bedroom} & \multicolumn{3}{c}{Living room} & \multicolumn{3}{c}{Dining room}
        \\ 
        \cmidrule(lr){2-4}
        \cmidrule(lr){5-7}
        \cmidrule(lr){8-10}
        &   FID & $\text{FID}_{\text{CLIP}}$ & KID & FID & $\text{FID}_{\text{CLIP}}$ & KID & FID & $\text{FID}_{\text{CLIP}}$ & KID \\
     \midrule 
        CommonScenes$^{*}$~\cite{zhai2024commonscenes} &  49.52 & \phantom{0}5.56 & \phantom{0}2.20 & 78.19 & \phantom{0}7.59 & \phantom{0}6.12 & 69.89 & \phantom{0}7.88 & \phantom{0}4.78 \\
        EchoScene$^{*}$~\cite{zhai2024echoscene} &  44.10 & \phantom{0}4.49 & \phantom{0}0.30 & 75.35  & \phantom{0}7.37 & \phantom{0}5.59 & 67.52 & \phantom{0}7.27 & \phantom{0}3.15 \\
        MMGDreamer$^{*}$\cite{yang2025mmgdreamer} &  42.38 & \phantom{0}4.48 & \phantom{0}-0.14 &  74.14 & \phantom{0}7.31 & \phantom{0}5.62 & 67.97 & \phantom{0}7.39 & \phantom{0}2.93 \\
     \textbf{FlowScene~(Ours)}  &  \textbf{35.01}& \phantom{0}\textbf{3.14} & \phantom{0}\textbf{-0.34} & \textbf{70.34}  & \phantom{0}\textbf{7.18} & \phantom{0}\textbf{5.11} & \textbf{60.26} & \phantom{0}\textbf{6.72} & \phantom{0}\textbf{0.07}  \\
    \midrule 
    CommonScenes~\cite{zhai2024commonscenes}  & 57.68 & \phantom{0}4.86 & \phantom{0}6.59 & 80.99 & \phantom{0}7.05 & \phantom{0}6.39 & 65.71 & \phantom{0}7.04 & \phantom{0}5.47\\
    EchoScene \cite{zhai2024echoscene}  &  {48.85} & \phantom{0}{4.26} & \phantom{0}{1.77} &  {75.95} & \phantom{0}{6.73} & \phantom{0}{0.60} & {62.85} & \phantom{0}{6.28} & \phantom{0}{1.72}\\
    MMGDreamer~\cite{yang2025mmgdreamer} & 45.75  & \phantom{0}3.84 & \phantom{0}{1.72} & {68.94}  & \phantom{0}{6.19} & \phantom{0}{0.40} & {55.17}  & \phantom{0}\textbf{5.86} & \phantom{0}0.05\\
    \textbf{FlowScene~(Ours, w/o TB)} &  \textbf{38.87} & \phantom{0}\textbf{3.56} & \phantom{0}\textbf{-0.54} & \textbf{67.83}  & \phantom{0}\textbf{6.11} & \phantom{0}\textbf{0.31} &  \textbf{54.38} & \phantom{0}5.94 & \phantom{0}\textbf{0.02}   \\
    \bottomrule
    \end{tabular}
    }
\label{tab:fidkid}
\end{table*}

\begin{table*}[t]
    \caption{\textbf{Object-Level Fidelity.} Quantitative evaluation of generated object quality and diversity using MMD ($\times0.01$), COV, and 1-NNA metrics across different categories. The best results are highlighted in \textbf{bold}.}
    \centering
    \scalebox{1}{
    \begin{tabular}{l ccccccccccc}
    \toprule
    Method & Metric & Bed & N.stand & Ward. & Chair & Table & Cabinet & Lamp & Shelf & Sofa & TV stand\\
    \midrule 
        CommonScenes~\cite{zhai2024commonscenes}& \multirow{4}{*}{MMD ($\downarrow$)} & {0.49} & {0.92} & {0.54} & 0.99 & {1.91} & {0.96} & {1.50} & {2.73} & {0.57} & {0.29} \\
        EchoScene~\cite{zhai2024echoscene} & & 0.37 & 0.75  & 0.39 & 0.62 & 1.47 & 0.83 & 0.66 & 2.52 & 0.48 & 0.35 \\
        MMGDreamer~\cite{yang2025mmgdreamer} & & {0.22} & {0.41} & {0.24}  & {0.35} & \textbf{0.55} & {0.71} & {0.34} & {1.58} & \textbf{0.43} & {0.24}  \\
        \textbf{FlowScene~(Ours)} & & \textbf{0.17} & \textbf{0.23} & \textbf{0.20} & \textbf{0.32}  & 0.64 & \textbf{0.60} & \textbf{0.20} & \textbf{1.53}  & 0.48 &\textbf{0.13}    \\
    \midrule 
        CommonScenes~\cite{zhai2024commonscenes}& \multirow{4}{*}{COV ($\%, \uparrow$)} & {24.07} & {24.17} & {26.62} & {26.72} & {40.52} & {28.45} & {36.21} & {40.00} & {28.45} & {33.62} \\
        EchoScene~\cite{zhai2024echoscene} & & 39.51 & 25.59 & 37.07 & 17.25 & 35.05 & 43.21 & 33.33 & 50.00 & 41.94 & 40.70\\
        MMGDreamer~\cite{yang2025mmgdreamer} & & {42.59} & {30.81}  & {44.44} & {19.95} & {44.12} & {49.38} & {40.56} &{70.00} & {47.31} &{45.35} \\
        \textbf{FlowScene~(Ours)} & &\textbf{69.54} & \textbf{56.46} &  \textbf{66.42} & \textbf{50.90} & \textbf{64.10} &\textbf{70.89} & \textbf{67.94} & \textbf{90.00} & \textbf{61.36} & \textbf{66.67}  \\
    \midrule 
        CommonScenes~\cite{zhai2024commonscenes}& \multirow{4}{*}{1-NNA ($\%, \downarrow)$} & {85.49} & {95.26} & {88.13} & {86.21} & {75.00} & {80.17} & {71.55} & {66.67} & {85.34} & {78.88} \\
        EchoScene~\cite{zhai2024echoscene} & & 72.84 & 91.00 & 81.90 & 92.67 & 75.74 & {69.14} & 78.90 & \textbf{35.00} & 69.35 & 78.49\\
        MMGDreamer~\cite{yang2025mmgdreamer} & & 69.44 &  {90.52} & {74.81} & {89.56} & {68.85} & {68.35} & {72.38} & {30.00} & {62.37} & {73.26} \\
        \textbf{FlowScene~(Ours)} & & \textbf{45.51} & \textbf{59.19} & \textbf{50.19}  & \textbf{79.60} & \textbf{54.83} & \textbf{59.12} & \textbf{53.22} & 10.00 & \textbf{52.25} & \textbf{54.44}  \\
    \bottomrule
    \end{tabular}
    }
    \label{tab:objfd}
\end{table*}

\subsection{Experimental Settings}
\paragraph{Datasets}
We conduct experiments on the SG-FRONT dataset~\cite{zhai2024commonscenes} and 3D-FRONT dataset~\cite{fu20213d}. SG-FRONT encompasses about 45K object instances and 15 categories of relationships across bedrooms, dining rooms, and living rooms. 
Following the protocol~\cite{yang2025mmgdreamer}, we obtain a multimodal graph where each node selectively includes textual and visual modalities.

\paragraph{Baselines}
We compare FlowScene to two categories of methods:
\emph{(i) Training-free, language-based methods for object retrieval}: Holodeck~\cite{yang2024holodeck} and LayoutVLM~\cite{sun2025layoutvlm}, which leverage large language and vision–language models~\cite{achiam2023gpt,li2024llava} to compose 3D scenes from text prompts. We obtain the prompts by translating scene graph labels to sentences.
\emph{(ii) Graph-conditioned generative models trained under the same protocol for both object retrieval and generation}: 
CommonScenes~\cite{zhai2024commonscenes}, which generates scenes from graphs using a VAE for layout and latent diffusion for shapes; 
EchoScene~\cite{zhai2024echoscene}, a dual-branch diffusion model for layout and shape; 
MMGDreamer~\cite{yang2025mmgdreamer}, which leverages a multimodal graph to emphasize geometry control;
and all models are trained under the same protocol.

\paragraph{Metrics}
We evaluate along four dimensions. (i) For \emph{generation realism}, we measure scene-level fidelity with Fréchet Inception Distance (FID)~\cite{heusel2017gans}, $\text{FID}_\text{CLIP}$~\cite{kynkaanniemi2022role}, and Kernel Inception Distance (KID)~\cite{binkowski2018demystifying} computed on top-down renderings of generation to measure similarity to ground truth, and object-level fidelity with Minimum Matching Distance (MMD), Coverage (COV), and 1-Nearest Neighbor Accuracy (1-NNA)~\cite{yang2019pointflow}. (ii) For \emph{generation controllability}, we report CLIPScore measuring the adherence between top-down renderings and user instructions~\cite{hessel2021clipscore}, FPVScore on multiple egocentric views~\cite{huang2025video}, and the graph-constraint satisfaction rate following prior work~\cite{zhai2024echoscene,yang2025mmgdreamer}. (iii) For \emph{style consistency}, we again adopt FPVScore and adapt its criteria to additionally focus on the geometric structure and visual appearance. (iv) Finally, we report the inference time for evaluating the \emph{generation efficiency} compared to other graph-based methods.

\begin{table*}[t!]
\caption{\textbf{Scene Perception Evaluation.} For FPVScore, we report the average ranking across all scenes, where 1.00 is the worst and 5.00 is the best. For the perceptual study, scores range from 1.00 to 10.00. Each score is averaged across all participants and scenes. Both FPVScore and the perceptual study benchmark prompt adherence (PA), layout correctness (LC), visual quality (VQ), style consistency (SC), and overall preference (OP). The best results are highlighted in \textbf{bold}. * denotes the retrieval mode.}
    \centering
    \scalebox{1}{
    \begin{tabular}{lccc|ccccc|ccccc}
        \toprule
        \textbf{Method} & \multicolumn{3}{c}{\textbf{CLIPScore$\uparrow$}} & \multicolumn{5}{|c|}{\textbf{FPVScore$\uparrow$}} & \multicolumn{5}{c}{\textbf{Perceptual Study}$\uparrow$} \\
        \cmidrule(lr){2-4} \cmidrule(lr){5-9} \cmidrule(lr){10-14}
        & Bedroom & Living room & Dining room & PA & LC & VQ & SC & OP & PA & LC & VQ & SC & OP  \\
        \midrule
        Echoscene$^{*}$~\cite{zhai2024echoscene} & 0.1837 & 0.1799 & 0.1553 &  3.03  & 3.07 & 2.82  & 3.03 & 3.01 &  7.21 &  6.83 & 6.86 & 7.12 &  7.24  \\

        MMGDreamer$^{*}$~\cite{yang2025mmgdreamer} & 0.1957 & 0.1877 & 0.1626  & 3.12 &  3.31  & 3.08  & 3.25 & 3.27 & 7.47 & 7.12 & 7.25 & 7.31 & 7.42 \\
        
        Holodeck~\cite{yang2024holodeck} & 0.1912  & 0.2116  & 0.1968
        &  2.79  &  2.73  &  2.71 & 2.70  &  2.70 & 7.16 & 6.61 & 7.43 & 6.72 & 6.64 \\
        LayoutVLM~\cite{sun2025layoutvlm} &  0.2045  &  0.2056  & 0.1947  &  2.36  & 2.44  & 2.15  & 2.23  & 2.31 & 6.52 & 6.33 & 7.21 & 6.43 & 6.32  \\

        \textbf{FlowScene~(Ours)} & \textbf{0.2386} &\textbf{0.2246} & \textbf{0.2092} & \textbf{3.35} & \textbf{3.54} & \textbf{3.83} & \textbf{3.85} & \textbf{3.76} & \textbf{7.97} & \textbf{7.65} & \textbf{8.46} & \textbf{8.72} & \textbf{8.57} \\
        \bottomrule
    \end{tabular}
    }
    \label{tab:scene_perception_evaluation}
\end{table*}

\paragraph{Perceptual Study}
We conduct a perceptual study based on ratings from 25 participants across 20 scenes to evaluate human preference alignment, including prompt adherence (PA), layout correctness (LC), visual quality (VQ), style consistency (SC), and overall preference (OP). Details are provided in~\autoref{userstudy}.

\paragraph{Implementation Details}
We perform all experiments on a single NVIDIA A100 GPU with 80 GB of memory.
The training of the three branches—shape, layout, and texture—is independently optimized using AdamW with an initial learning rate of 1e-4 and a batch size of 196. 
The shape and layout branches use a flow transformer, while the texture branch employs a sparse flow transformer, both following~\cite{xiang2024structured}. The sampling step $K$ is 25.
More implementation details are provided in~\autoref{imp_detail}.


\begin{figure*}[h]
  \centering
  \vspace{15px}
  \begin{overpic}[width=0.95\textwidth,trim=0cm 0 0 0cm,clip]{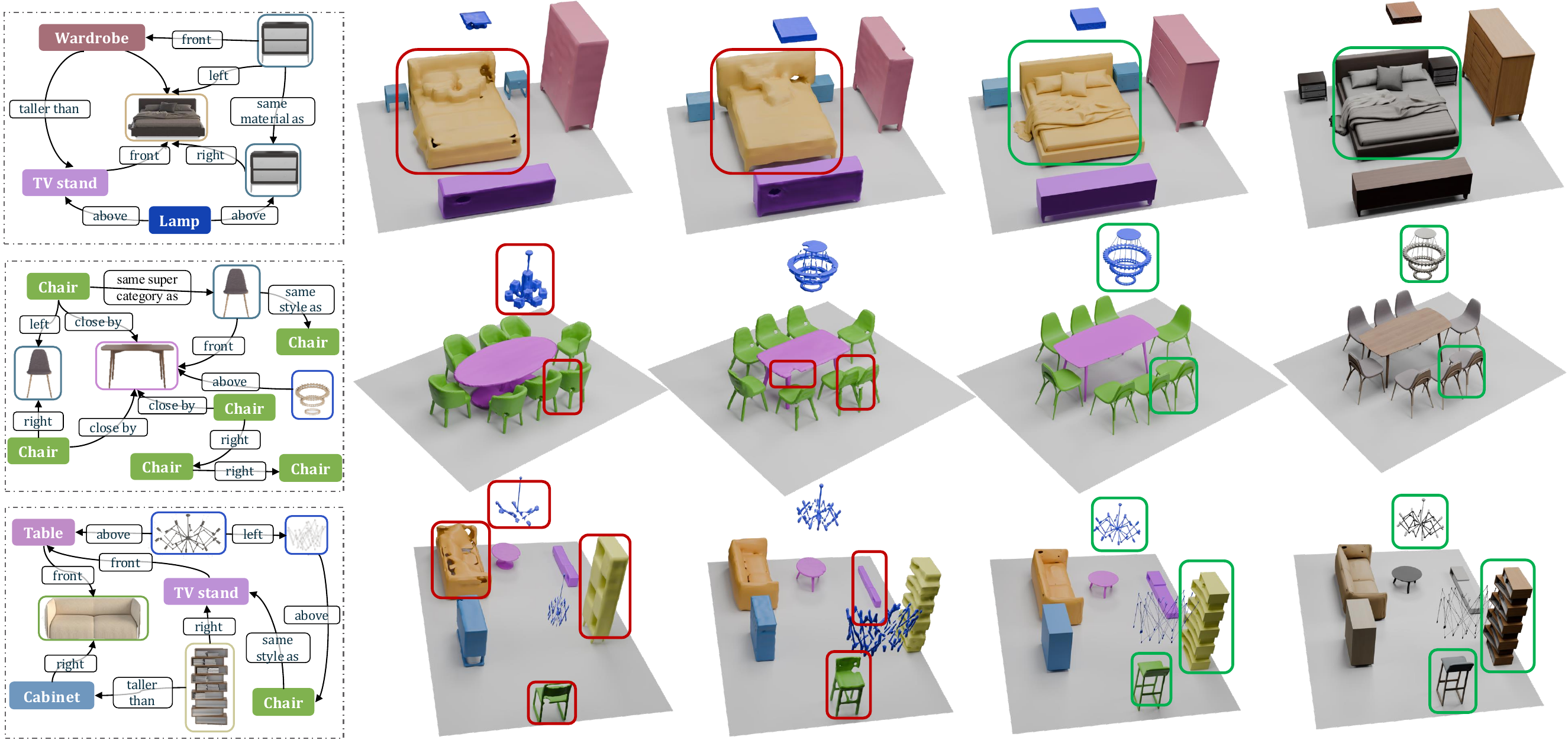}
    \put(6.5,48.3){\color{black}{\small Input Graph}}
    \put(27,48.3){\color{black}{\small EchoScene$^\dag$}~\cite{zhai2024echoscene}}
    \put(45,48.3){\color{black}{\small MMGDreamer}~\cite{yang2025mmgdreamer}}
    \put(62.5,48.3){\color{black}{\small {FlowScene w/o TB}}}
    \put(85,48.3){\color{black}{\small \textbf{FlowScene}}}

    \put(-2.7,35.5){\rotatebox{90}{\small Bedroom}}
    \put(-2.7,18){\rotatebox{90}{\small Dining Room}}
    \put(-2.7,3){\rotatebox{90}{\small Living Room}}
    
  \end{overpic}
    \caption{\textbf{Qualitative comparison.} The first column shows the input multimodal graph with key scene edges. Red rectangles show inconsistencies and artifacts, while green rectangles highlight consistent regions. $^\dag$Text-only scene graphs are given.
}
    \label{demo}
\end{figure*}

\begin{figure*}[t]
  \centering
  \begin{overpic}[width=1\linewidth,trim=0cm 0 0 0cm,clip]{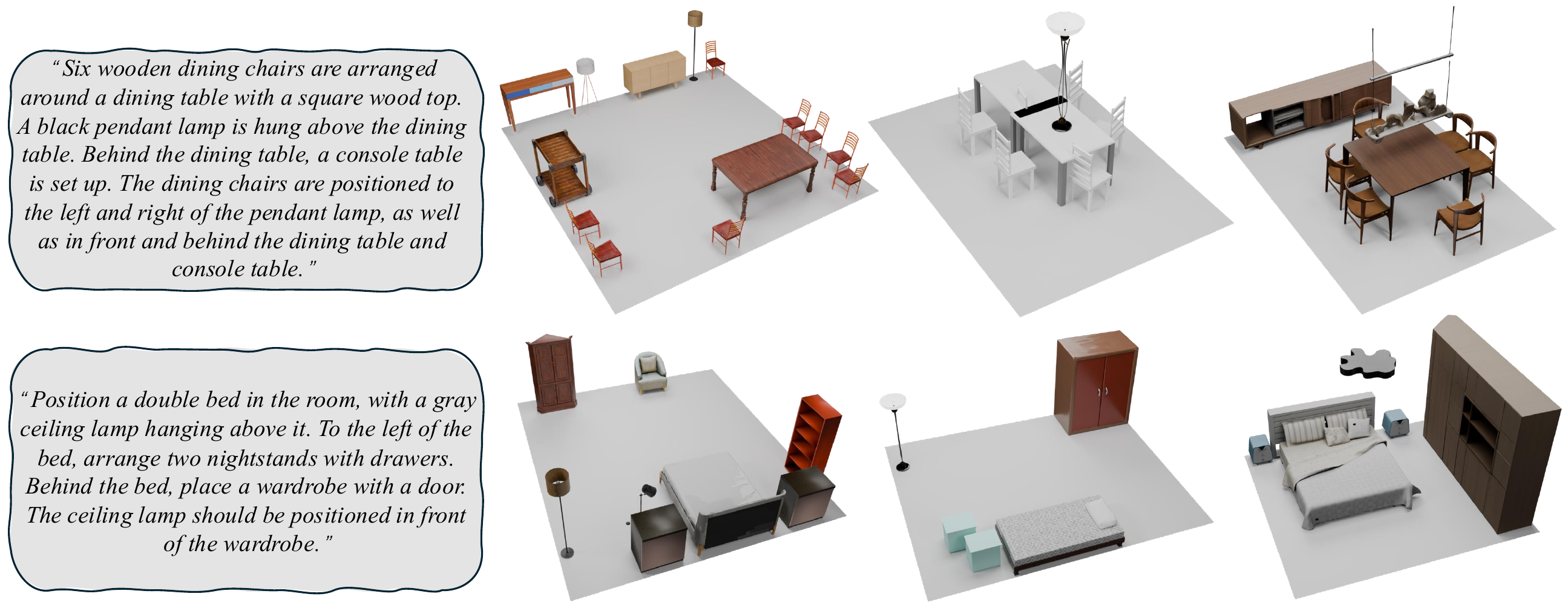}
  \put(11,-1){\color{black}{Instructions}}
    \put(40,-1){\color{black}{Holodeck~\cite{yang2024holodeck}}}
    \put(63,-1){\color{black}{LayoutVLM~\cite{sun2025layoutvlm}}}
    \put(85,-1){\color{black}{\textbf{FlowScene~(Ours)}}}
  
  \end{overpic}
    \caption{\small \textbf{Qualitative comparison} with training-free, language-based methods. The grey boxes display the input textual instructions derived from text-only scene graphs, while FlowScene takes the text-only graphs as input.
}
    \label{demo_llm}
\end{figure*}

\subsection{Quantitative results}

\paragraph{Scene-Level Realism}
We compare FlowScene with graph-based baselines~\cite{zhai2024commonscenes,zhai2024echoscene,yang2025mmgdreamer} in two settings: First, baselines operate in retrieval mode, composing textured objects into 3D scenes, and we evaluate FlowScene with its texture branch. Second, all methods generate both layouts and shapes without textures, as baselines are not able to produce textures. This fairly assesses holistic geometric fidelity only.
As shown in~\autoref{tab:fidkid}, FlowScene consistently outperforms all methods in the retrieval setting, even though others retrieve well-designed meshes. For bedroom in particular, FlowScene improves over MMGDreamer by reducing FID by 7.37, $\text{FID}_\text{CLIP}$ by 1.34, and KID by 0.20, which indicates a closer match to the real-scene distribution. A similar pattern appears in the full generation setting, where FlowScene achieves the lowest FID and KID among all baselines. In the perceptual study, as shown in~\autoref{tab:scene_perception_evaluation}, LC and VQ also indicate that FlowScene exhibits the best alignment with human preference.


\paragraph{Object-Level Realism}
Object-level metrics focus on fine-grained geometric accuracy and diversity rather than the holistic scene fidelity. As shown in~\autoref{tab:objfd}, FlowScene achieves better MMD and COV across most object categories. Specifically, for nightstands and lamps, MMD decreases by 43.90\% and 41.76\%, while COV increases by 45.43\% and 40.30\%, respectively, over MMGDreamer~\cite{yang2025mmgdreamer}. The 1-NNA results further show that FlowScene matches the shape distribution more closely than competing methods. Overall, these results demonstrate the superior realism and generative quality of objects produced by FlowScene.

\paragraph{Generation Controllability}
Controllability measures how well generated scenes follow the given conditions or instructions. 
Since all methods except MMGDreamer and FlowScene take only textual inputs, we construct text-only scene graphs for all methods to ensure a fair comparison and evaluate vision–text consistency with CLIPScore~\cite{hessel2021clipscore}  and FPVScore~\cite{huang2025video}. 
CLIPScore is computed on top-down renderings against a single coarse sentence that encodes all semantic and spatial information, which is widely used but relatively rough. FPVScore is a more recent approach that utilizes egocentric views with detailed prompts to provide fine-grained measurements of spatial and semantic adherence. As shown in~\autoref{tab:scene_perception_evaluation}, FlowScene attains the highest CLIPScore across all room types, and outperforms all baselines on the PA of FPVScore and perceptual study, indicating strong alignment with textual descriptions. 
We additionally provide the measurements of graph constraints in~\autoref{addition}.

\paragraph{Style Consistency}

Style consistency is harder to quantify than fidelity or controllability, since it depends on subjective judgments about whether objects share a consistent geometry and appearance within a scene. To obtain a more reliable and comprehensive measurement, we extend FPVScore with a new criterion that focuses specifically on object structure and visual appearance, as illustrated in~\autoref{fpv_detail}. Benefiting from multiple continuous views covering the whole scene, this criterion is designed to more effectively capture whether objects match the intended style at both the geometric and textural levels, rather than only matching semantics. As shown in~\autoref{tab:scene_perception_evaluation}, FlowScene outperforms all baselines on both VQ and SC, reflecting its strong ability to generate scenes that are both high-quality and consistent in cross-object style. The perceptual study on SC leads to the same conclusion, further supporting the effectiveness of our style modeling.

\paragraph{Generation Efficiency}
We measure the inference speed for evaluating efficiency, averaging over 10 scenes. All methods are conducted on a single NVIDIA A100 GPU. As shown in~\autoref{tab:inference_time}, FlowScene demonstrates significantly faster speed in parallelly generating layout and shape, achieving an inference time of 6.83 seconds, which is 84.93\% faster than MMGDreamer. Even though the texture branch is enabled, FlowScene still maintains the fastest inference time, with a result of 37.38 seconds.
\begin{table}[t]
\centering
\caption{\textbf{Inference Time Comparison.}}
\label{tab:inference_time}
\scalebox{0.92}{
\begin{tabular}{lcccccc}
\toprule
Method & Layout & Shape & Texture & Time~(s) \\
\midrule
EchoScene~\cite{zhai2024echoscene} & \cmark & \cmark &  -- & 44.27 \\
MMGDreamer~\cite{yang2025mmgdreamer} & \cmark & \cmark & -- & 45.34 \\
\midrule
{\textbf{FlowScene~(Ours, w/o TB)}} & \cmark & \cmark & \xmark & 6.83 \\
{\textbf{FlowScene~(Ours)}} & \cmark & \cmark & \cmark & 37.38 \\
\bottomrule
\end{tabular}
}
\end{table}

\begin{table}[t]
\centering
\caption{\textbf{Ablation Study on the InfoExchangeUnit} across different branches, including the \textsc{LayoutExchangeUnit}~(LEU), \textsc{ShapeExchangeUnit}~(SEU), and \textsc{TextureExchangeUnit}~(TEU).}
\scalebox{0.92}{
    \begin{tabular}{cccccc}
    \toprule
        LEU &SEU & TEU & FID & $\text{FID}_\text{CLIP}$ & KID \\
     \midrule
         &&  & 50.83  & 4.55 & 7.32  \\
         \cmark &&  & 40.55 & 3.48 & 2.58 \\
          & \cmark & &44.73 & 4.42  & 4.97   \\
          &  & \cmark & 43.82  &3.86  &3.68  \\
         \cmark & \cmark & & 36.81 & 2.92  & 1.31   \\
         \cmark &  & \cmark & 36.30  & 2.90 & 0.77 \\
          & \cmark & \cmark & 39.35  & 3.22 & 2.08 \\
    \midrule 
       \cmark & \cmark & \cmark  & {32.76} &{2.86} & {0.38}\\
     \bottomrule
    \end{tabular}
    }
\label{tab:ablation_IEU}
\end{table}

\begin{table}[t]
\caption{\textbf{Ablation studies} on different generation backbones.} 
\centering
\scalebox{0.92}{
    \begin{tabular}{ccccccc}
    \toprule
            \multicolumn{2}{c}{Shape} & \multicolumn{2}{c}{Layout}& \multirow{2}{*}{{FID}} &\multirow{2}{*}{$\text{FID}_\text{CLIP}$} & \multirow{2}{*}{{KID}}  \\
    \cmidrule(lr){1-2} \cmidrule(lr){3-4}
           Diffusion & Flow & Diffusion & Flow &  \\
     \midrule
           \cmark & & \cmark &  & 35.23 & 3.26 & 1.78  \\
           \cmark &  &  & \cmark & 34.96 & 2.90 & 1.03 \\
    
            &  \cmark &\cmark &  & 33.03 & 2.94 & 0.62 \\
          &  \cmark & & \cmark & \textbf{32.76} & \textbf{2.86} & \textbf{0.38}  \\
     \bottomrule
    \end{tabular}
    }
\label{tab:ablation_model}
\end{table}

\begin{table}[t]
\caption{\textbf{Ablation studies} on the ratio of node modalities.} 
\centering
\scalebox{0.92}{
\begin{tabular}{cccccc}
\toprule
\multicolumn{2}{c}{\textbf{Modality}} & \multicolumn{4}{c}{\textbf{FPVScore$\uparrow$}} \\
\cmidrule(lr){1-2} \cmidrule(lr){3-6}
Visual & Textual & LC & VQ & SC & OP \\
\midrule
10\% & 90\% & 3.28 & 3.37 & 3.42 & 3.26  \\
40\% & 60\% & 3.45 & 3.86 & 3.89 & 3.67  \\
50\% & 50\% & 3.42 & 3.88 & 3.91 & 3.64  \\
60\% & 40\% & 3.36 & 3.88 & 3.91 & 3.62  \\
90\% & 10\% & 3.28 & 3.78 & 3.82 & 3.59  \\
\bottomrule
\end{tabular}
}
\label{tab:ablation_ratio}
\end{table}

\subsection{Qualitative results}
We perform qualitative evaluation across different room types, as shown in~\autoref{demo}. FlowScene demonstrates superior style consistency, particularly in the dining room, where the generated chairs maintain a consistent appearance and texture, even as text-only nodes, while other methods fail to generate textures. 
This emphasizes FlowScene's ability to generate both geometrically accurate and visually consistent scenes, outperforming baseline methods like MMGDreamer. 
Furthermore,~\autoref{demo_llm} highlights FlowScene's improved controllability compared with the language-driven methods Holodeck~\cite{yang2024holodeck} and LayoutVLM~\cite{sun2025layoutvlm}. To achieve a fair comparison, we condition FlowScene on the text-only scene graphs. 
FlowScene enables more controllable instruction-guided generation, whereas baselines either exhibit style inconsistencies or fail to satisfy layout constraints accurately.
This confirms that FlowScene not only enforces holistic coherence across multiple objects but also bridges abstract language with precise 3D generation, surpassing graph-condition or pure language-based approaches.
We provide additional qualitative results in~\autoref{qualitative}.

\subsection{Ablation Study}
\paragraph{Comparison on InfoExchangeUnit}
As shown in~\autoref{tab:ablation_IEU}, we ablate different combinations of \textsc{LayoutExchangeUnit}~(LEU), \textsc{ShapeExchangeUnit}~(SEU), and \textsc{TextureExchangeUnit}~(TEU).
Without any exchange unit, the model obtains an FID of 50.83. Enabling a single exchange unit consistently improves performance, with LEU giving the largest gain, highlighting the importance of layout-level information exchange for spatial coherence. Combining exchange units further improves results, especially LEU with TEU, which reduces KID to 0.77 and indicates better appearance consistency across objects. The best performance is achieved when all three units are enabled, yielding the lowest FID, FID$_{\text{CLIP}}$, and KID.

\paragraph{Comparison on Different Backbones}
We steadily replace the diffusion backbone in the shape and layout branches with our graph flow. As shown in~\autoref{tab:ablation_model}, switching only the layout branch from diffusion to flow brings a modest gain in scene fidelity. 
In contrast, switching only the shape branch yields a larger improvement, with FID dropping from 35.23 to 33.03.
Using graph flow in both branches yields the best results, indicating that most of the benefit comes from the shape branch, while the layout branch provides complementary gains.

\paragraph{Comparison on Different Modality Ratios}
We evaluate how robust style consistency and visual quality are to different ratios of visual and textual information in the multimodal graph. Starting from 10\% of nodes with visual inputs, we gradually increase the ratio to 90\%. As shown in~\autoref{tab:ablation_ratio}, style consistency, measured by SC, remains stable at 3.79 with a variance of 0.03, and the accuracy-related metrics LC, VQ, and OP also stay stable across all ratios.

\section{Discussion}
\label{sec:dis}
Experiments have shown that FlowScene serves as a robust scene generator driven by multimodal input. To broaden its use, we design an application interface in~\autoref{interface} that illustrates how FlowScene supports scene creation from diverse input sources and interactive user operations. While our current experiments are conducted on a single dataset, the scalability of FlowScene to larger and more diverse environments remains to be further explored, this does not diminish the generality and promise of the proposed interaction framework. Future directions include extending FlowScene to larger and outdoor scene, and integrating more deeply with interactive design and planning tools.

\section{Conclusion}
\label{sec:con}
We introduced \emph{FlowScene}, a tri-branch generator conditioned on multimodal graphs that generates scene layouts, object shapes, and object textures while preserving scene-level style consistency across structure and appearance. 
Each branch is backboned by a  \emph{Multimodal Graph Rectified Flow} module that exchanges object information during sampling, which enables fine-grained control over per-object appearance and relations. This module also improves efficiency compared with diffusion-based pipelines. 
Experiments show consistent gains in realism, controllability, style consistency, and human preference.

\bibliographystyle{ACM-Reference-Format}
\bibliography{sample-base}

\appendix
\clearpage

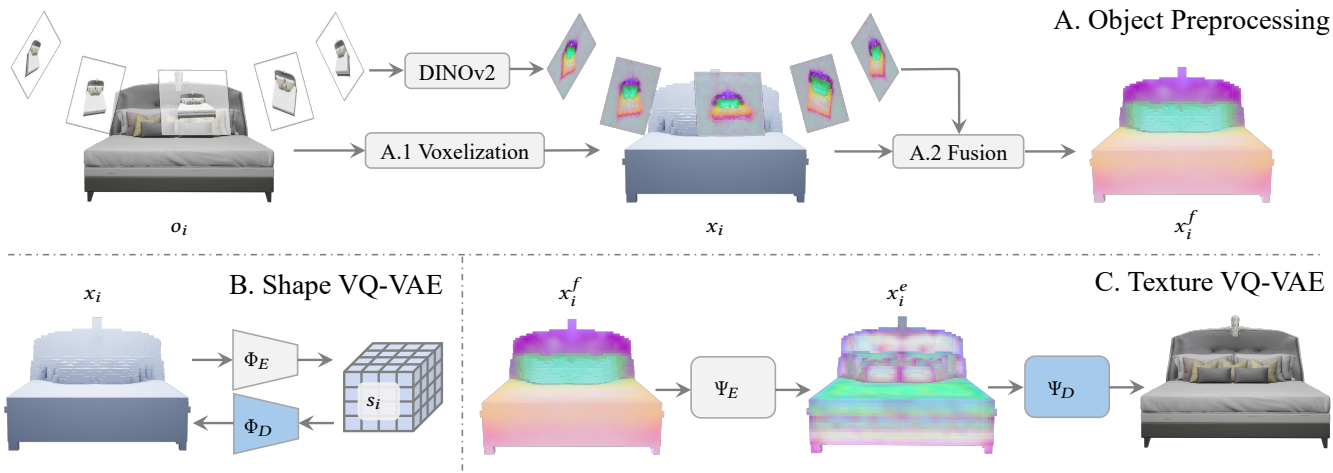
\begin{figure*}
\centering
\vspace{6pt}
\begin{overpic}[width=\textwidth,trim=0 0 0 0,clip]{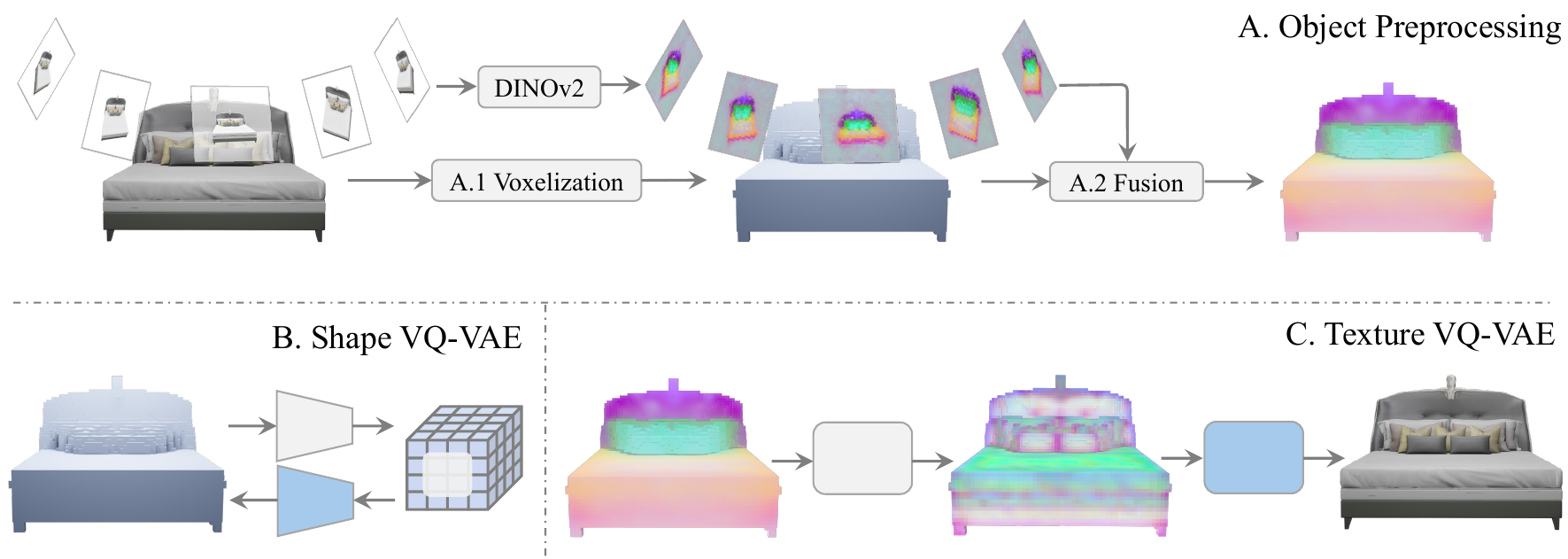}
  \put(13,18){\color{black}{\small $o_i$}}
  \put(53,18){\color{black}{\small $x_i$}}
  \put(88,18){\color{black}{\small $x_i^f$}}

  \put(6.5,13){\color{black}{\small $x_i$}}
  \put(42,13){\color{black}{\small $x_i^f$}}
  \put(66.5,13){\color{black}{\small $x_i^e$}}
  \put(18.5,8){\color{black}{\small $\Phi_{E}$}}
  \put(18.5,3){\color{black}{\small $\Phi_{D}$}}
  \put(27.6,5){\color{black}{\small $s_i$}}

  \put(53.6,5.8){\color{black}{\small $\Psi_{E}$}}
  \put(78.5,5.8){\color{black}{\small $\Psi_{D}$}}
\end{overpic}

\captionof{figure}{\textbf{Data Preparation.}}
\label{fig:process}
\end{figure*}

\paragraph{Supplementary Material Overview.}
\begin{itemize}
  \setlength{\itemsep}{2pt}
  \setlength{\parskip}{0pt}
  \setlength{\parsep}{0pt}

  \item Object preprocessing pipeline (\autoref{sec:object})
  \item Additional experimental results (\autoref{addition})
  \item Qualitative comparisons (\autoref{qualitative})
  \item Implementation details (\autoref{imp_detail})
  \item FPVScore criteria and perceptual study protocol (\autoref{fpv_detail}, \autoref{userstudy})
  \item Applications: language-driven and GUI-driven interfaces (\autoref{interface})
  \item Limitations and failure cases (\autoref{limitation})
\end{itemize}

\section{Object Processing}
\label{sec:object}

We show the object processing stages in~\autoref{fig:process}. Object $o_i$ is voxelized into $x_i$ (\autoref{fig:process} \textcolor{cvprblue}{A.1}), on which we train $\Phi_{(E,D)}$ to obtain the latent shape code $s_i$ (\autoref{fig:process} \textcolor{cvprblue}{B}). In parallel, we render multi-view images of $o_i$ and extract DINOv2 features, which are reprojected onto $x_i$ and averaged across views to yield feature voxels $x_i^f$ (\autoref{fig:process} \textcolor{cvprblue}{A.2}). We then train $\Psi_{(E,D)}$, encoding $x_i^f$ into $x_i^e$ and decoding it to the original object $o_i$ (\autoref{fig:process} \textcolor{cvprblue}{C}).

\section{Additional Results}
\label{addition}

\paragraph{Scene Graph Constraints}
\begin{table*}[t!]
    \caption{\textbf{Scene graph constraints.} Comparisons are across three modes: Relationship Change, Node Addition, and Generation Only (listed top to bottom). The best results are highlighted in \textbf{bold}.}
    \centering
    \scalebox{0.90}{
    \begin{tabular}{l c cccc cc}
    \toprule
    \multirow{2}{*}{Method}  & \multirow{2}{*}{\makecell{Mode}} & \multirow{2}{*}{\makecell{left/\\right}} & \multirow{2}{*}{\makecell{front/\\behind}} & \multirow{2}{*}{\makecell{smaller/\\larger}} & \multirow{2}{*}{\makecell{taller/\\shorter}} & \multirow{2}{*}{\makecell{close by}} & \multirow{2}{*}{\makecell{symmetrical}} \\
    &  & & & & & &\\
        
    \midrule 
        Graph-to-3D~\cite{dhamo2021graph}& \multirow{5}{*}{Change} & 0.91 & 0.92 & 0.86 & 0.89 & 0.69 & 0.46 \\
        CommonScenes~\cite{zhai2024commonscenes} &  & 0.91 & 0.92 & 0.86 & 0.91 & 0.69 & \textbf{0.59} \\
         EchoScene~\cite{zhai2024echoscene}& & {0.94} & \textbf{0.96} & {0.92} & {0.93} & {0.74} & 0.50\\
         MMGDreamer~\cite{yang2025mmgdreamer}  && \textbf{0.95}& \textbf{0.96} & \textbf{0.93} & {0.93} & 0.71 & {0.53}  \\
        \textbf{FlowScene~(Ours)} && \textbf{0.95} & \textbf{0.96} & \textbf{0.93} & \textbf{0.94} &\textbf{0.75}  &0.58   \\
        \midrule
        Graph-to-3D~\cite{dhamo2021graph} & \multirow{5}{*}{Addition} &  0.94 & 0.95 & 0.91 & 0.93 & 0.63 & 0.47\\
        CommonScenes~\cite{zhai2024commonscenes} & & 0.95 & 0.95 & 0.91 & 0.95 & 0.70 & {0.61}\\
         EchoScene~\cite{zhai2024echoscene} & & \textbf{0.98} & {0.99} & {0.96} & {0.96} & {0.76} & 0.49\\
         MMGDreamer~\cite{yang2025mmgdreamer} & & \textbf{0.98} & \textbf{1.00} & \textbf{0.97} & {0.97} & \textbf{0.80} & {0.61} \\
         \textbf{FlowScene~(Ours)} &&  \textbf{0.98}& \textbf{1.00} &\textbf{0.97} & \textbf{0.98} &0.77  & \textbf{0.68} \\
    \midrule
        Graph-to-3D~\cite{dhamo2021graph} & \multirow{5}{*}{None} & \textbf{0.98} & 0.99 & \textbf{0.97} & 0.95 & 0.74 & 0.57\\
        CommonScenes \cite{zhai2024commonscenes} & & \textbf{0.98} & \textbf{1.00} & \textbf{0.97} & 0.95 & \textbf{0.77} & {0.60} \\
        EchoScene~\cite{zhai2024echoscene}  & & \textbf{0.98} & 0.99 & {0.96} & {0.96} & 0.74 & 0.55\\
        MMGDreamer~\cite{yang2025mmgdreamer} & & \textbf{0.98} & {0.99} & \textbf{0.97} &{0.96}  &{0.76}  & \textbf{0.62} \\
        \textbf{FlowScene~(Ours)} && \textbf{0.98}& \textbf{1.00} & \textbf{0.97}& \textbf{0.97} & 0.76 & 0.58  \\
    \bottomrule
    \end{tabular}}
    \label{tab:sgconst}
\end{table*}
\begin{table*}[t!]
\caption{\textbf{Inter-shape consistency.} The Chamfer Distance (CD, $\times 0.001$) is computed between object pairs linked by the same as relationship. Lower values indicate a higher degree of shape similarity between paired instances.}
\centering
\scalebox{0.95}{
\begin{tabular}{l cc cc cc}
    \toprule
         \multirowcell{2}{Method} & \multicolumn{2}{c}{Bedroom} & \multicolumn{2}{c}{Living room} & \multicolumn{2}{c}{Dining room}\\
         \cmidrule{2-7}
          & N.stand &Lamp & Table &Lamp & Chair & Table\\
     \midrule
         CommonScenes \cite{zhai2024commonscenes} &2.69 &- &11.75 & - & 1.96 &  9.04  \\
         EchoScene \cite{zhai2024echoscene}  & 1.68& 30.07& 3.02 & 10.06&1.75  & 1.26 \\
         MMGDreamer\cite{yang2025mmgdreamer}  & {1.33}& {2.29}& {2.44} & {0.18} &{0.23} &{0.83} \\
         \textbf{FlowScene~(Ours)}  & \textbf{0.46} & \textbf{0.19} & \textbf{1.61} & \textbf{0.14} &\textbf{0.21}  & \textbf{0.37}  \\
     \bottomrule
    \end{tabular}
    }
\label{tab:consistency1}
\end{table*}

\begin{table}[t]
\caption{\textbf{Ablation studies of different input views.}} 
\centering
\scalebox{1}{
    \begin{tabular}{ccccccc}
    \toprule
            \multicolumn{2}{c}{Training} & \multicolumn{2}{c}{Testing}& \multirow{2}{*}{{FID}} &\multirow{2}{*}{$\text{FID}_\text{CLIP}$} & \multirow{2}{*}{{KID}} \\
            \cmidrule(lr){1-2} \cmidrule(lr){3-4}
           Multi & Single & Multi & Single &  \\
     \midrule
           \cmark & & \cmark &  & 33.77 & 2.88 &0.73   \\
           \cmark &  &  & \cmark & 33.79 & 2.97 &0.99  \\
            &  \cmark &\cmark & & 40.73 & 4.04 & 3.79\\
          &  \cmark & & \cmark & {32.76} &{2.86} & {0.38} \\
     \bottomrule
    \end{tabular}
    }
\label{tab:ablation_view}
\end{table}

\begin{table}[t]
\centering
\caption{\textbf{Ablation Study on the mask ratio of relation.} * indicates results from MMGDreamer.}
\scalebox{1}{
    \begin{tabular}{cccc}
    \toprule
         Mask Ratio & FID & $\text{FID}_\text{CLIP}$ & KID \\
     \midrule
         20\%   & 34.40 & 2.83 & 0.43 \\
         50\%   & 38.94 & 3.22 & 0.96 \\
         80\%   & 43.54 & 3.78 & 2.19 \\
    \midrule
         0\%$^{*}$ & 40.89 & 3.36& 2.05  \\
     \bottomrule
    \end{tabular}
    }
\label{tab:ablation}
\end{table}

Following previous graph-based works~\cite{zhai2024commonscenes,zhai2024echoscene,yang2025mmgdreamer}, we assess the model's ability to maintain semantic consistency between the input graph and the generated 3D scene across three distinct modes: Relationship Change, Node Addition, and Generation Only.
As shown in~\autoref{tab:sgconst}, even though the results are close to being saturated, FlowScene still achieves the best performances in the evaluation of most constraints, and exhibits robust performance in preserving spatial consistency during edge manipulations. For example, in the Relationship Change mode, FlowScene effectively adjusts object positions to satisfy updated constraints, achieving the best performance on complex relations, such as \texttt{close by}. Similarly, in the Node Addition mode, FlowScene demonstrates precise context-aware placement for newly added objects. It outperforms MMGDreamer on the challenging \texttt{symmetrical} metric. This indicates that our graph rectified flow mechanism can flexibly propagate local updates to the holistic scene structure.

\paragraph{Inter-Shape Consistency}
To quantify the geometric coherence of identical object shapes, the Chamfer Distance (CD) is computed between pairs linked by the \texttt{same as} relationship, where lower scores denote higher shape consistency.
As shown in~\autoref{tab:consistency1}, FlowScene achieves the state-of-the-art performance, consistently yielding the lowest CD values across all room types and object categories.
In the Bedroom scenario, a striking improvement is observed in the ``Lamp" category. Previous diffusion-based methods, such as EchoScene, struggle significantly, and FlowScene achieves a remarkable CD, surpassing MMGDreamer by an order of magnitude in the meantime. This demonstrates exceptional stability in generating consistent fine-grained structures.
A similar trend persists in the Living Room, where our model reduces the geometric discrepancy for ``Tables" and significantly outperforms CommonScenes.
Furthermore, in the Dining Room, FlowScene achieves the lowest CD for chairs, demonstrating that it successfully synchronizes geometric latents across spatially distinct nodes to ensure holistic consistency.

\paragraph{Ablation on Different Input Views}
We investigate the impact of visual input variability by training FlowScene with either single-view or multi-view images.
As detailed in~\autoref{tab:ablation_view}, while the model trained strictly on single-view images achieves the lowest FID under the matched testing condition, it fails to generalize to multi-view inputs, suffering a significant performance drop (FID increases by 7.97). In contrast, training with multi-view images yields remarkable robustness. The model exhibits negligible performance variance when switching from multi-view to single-view testing. This suggests that leveraging multi-view images during training encourages the model to learn view-invariant 3D representations, enabling it to adapt flexibly to varying visual conditions at inference without retraining.

\paragraph{Ablation on the Mask Ratio of Relations}
We conduct relation masking on the scene graphs of the SG-Front dataset~\cite{zhai2024commonscenes} to analyze how graph relation density affects scene generation fidelity, as summarized in~\autoref{tab:ablation_ratio}. As the mask ratio decreases, performance consistently improves, indicating that denser inter-object relations provide stronger structural guidance for scene generation. Notably, even with a moderate mask ratio of 50\%, FlowScene still outperforms the baseline without relation masking (0\%$^*$ from MMGDreamer~\cite{yang2025mmgdreamer}), demonstrating robustness to partial relation removal and the effectiveness of leveraging even sparse relational information.
Overall, these results highlight the importance of inter-object relations in enabling the graph-rectified flow to effectively propagate contextual information and enforce scene-level consistency across both structure and appearance during scene generation.

\section{Qualitative Results}
\label{qualitative}
\begin{figure*}[t]
  \centering
  \begin{overpic}[width=0.88\linewidth,trim=0cm 0 0 0cm,clip]{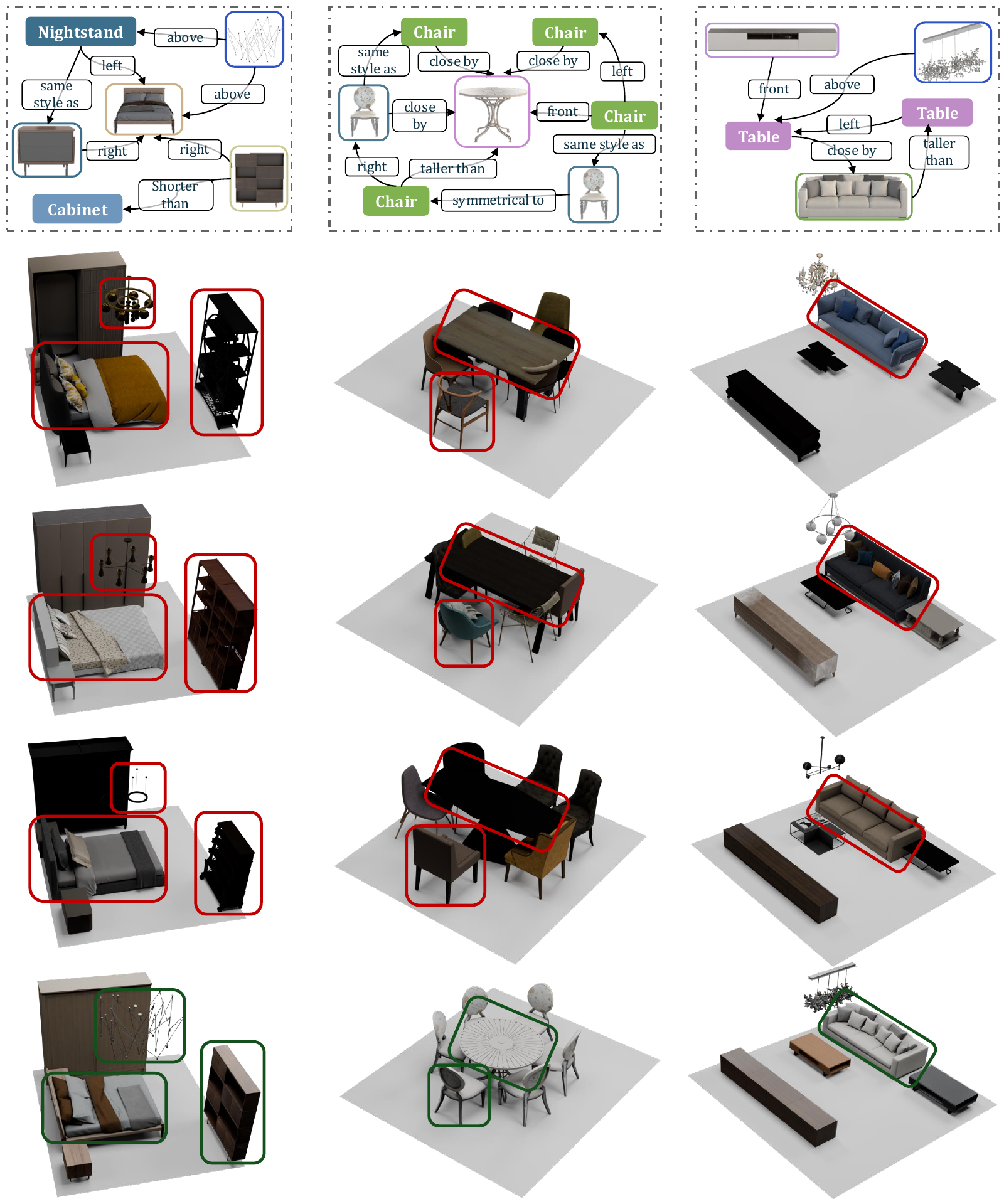}
    \put(8.7,100.5){\color{black}{Bedroom}}
    \put(36,100.5){\color{black}{Dining Room}}
    \put(65,100.5){\color{black}{Living Room}}

    \put(-2.5,61){\rotatebox{90}{ CommonScenes$^{\textbf{*,\dag}}$~\cite{zhai2024commonscenes}}}
    \put(-2.5,43){\rotatebox{90}{ EchoScene$^{\textbf{*,\dag}}$~\cite{zhai2024echoscene}}}
    \put(-2.5,23){\rotatebox{90}{ MMGDreamer$^{\textbf{*}}$~\cite{yang2025mmgdreamer}}}
    \put(-2.5,3){\rotatebox{90}{ \textbf{FlowScene~(Ours)}}}
  
  \end{overpic}
    \caption{\small \textbf{Qualitative comparison} with graph-conditioned generative models. The top row illustrates the input graphs, highlighting key semantic and spatial edges. In the generated scenes, red rectangles indicate generation failures in baselines (e.g., collisions), whereas green rectangles mark consistent regions in our results. * denotes the retrieval mode. $^\dag$Text-only scene graphs are given.
}
    \label{demo_texture}
\end{figure*}

\paragraph{Scene-Level Generation}
We conduct a qualitative comparison to demonstrate FlowScene's superiority in both consistency and controllability. As shown in~\autoref{demo_texture}, compared to graph-conditioned baselines, our method excels in style consistency and layout plausibility. For instance, in the Dining Room scenario, while baselines often produce disparate object styles or chaotic placements, FlowScene generates a set of chairs with unified appearance and texture, faithfully adhering to satisfying \texttt{the same style as} constraint.

\begin{figure*}[h]
  \centering
  \vspace{6px}
  \begin{overpic}[width=0.87\linewidth,trim=0cm 0 0 0cm,clip]{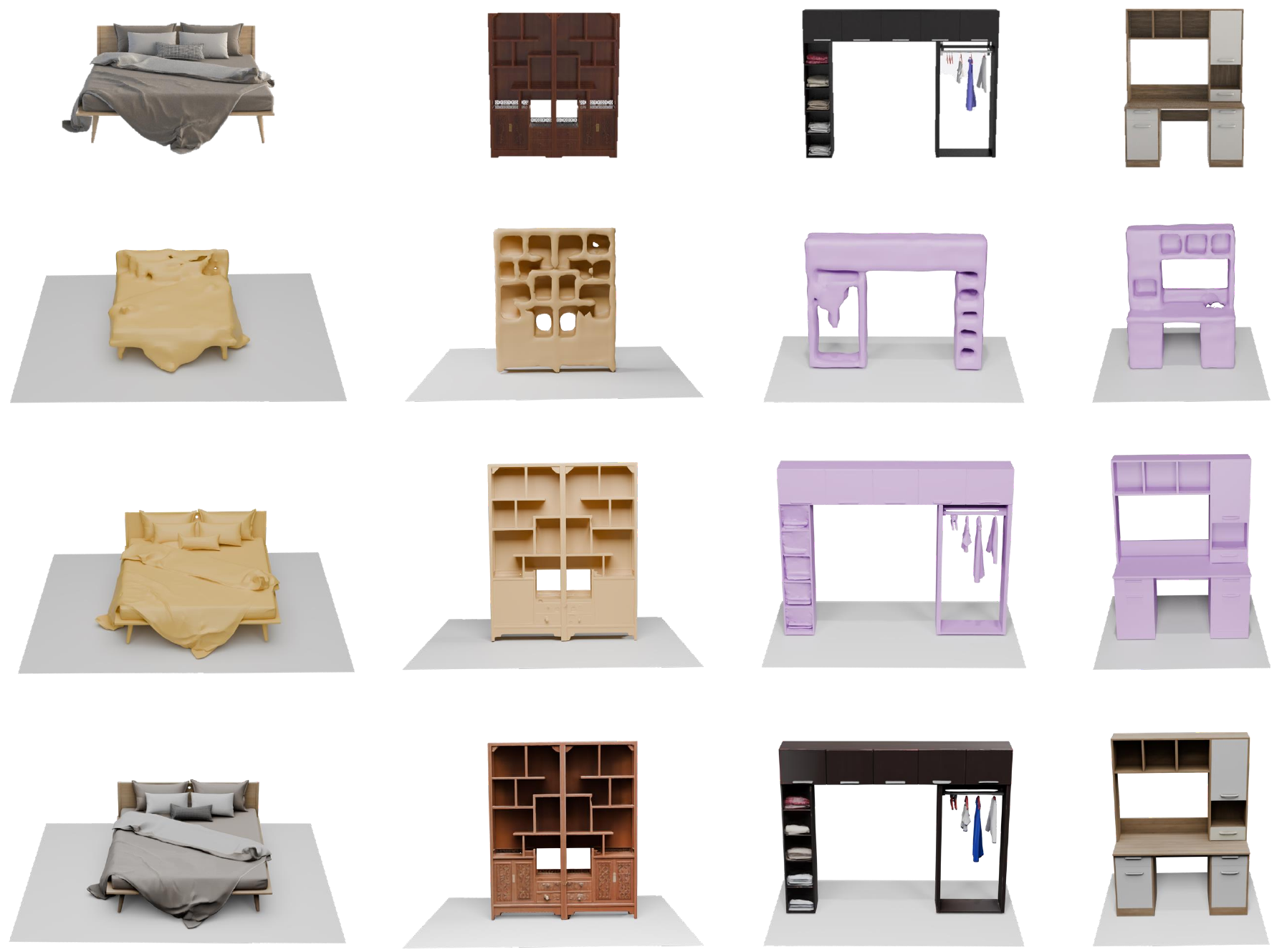}
    \put(11.5,74){\color{black}{\small Bed}}
    \put(38.5,74){\color{black}{\small Bookshelf}}
    \put(65.5,74){\color{black}{\small Wardrobe}}
    \put(89.5,74){\color{black}{\small Desk}}
    
    \put(-2.5,63){\rotatebox{90}{\small Node Image}}
    \put(-2.5,42){\rotatebox{90}{\small MMGDreamer~\cite{yang2025mmgdreamer}}}
    \put(-2.5,21){\rotatebox{90}{\small FlowScene~(w/o TB)}}
    \put(-2.5,0){\rotatebox{90}{\small \textbf{FlowScene~(Ours)}}}
  
  \end{overpic}
    \caption{\small \textbf{Qualitative comparison} on object generation with other methods.
}
    \label{demo_object}
\end{figure*}

\paragraph{Object-Level Generation}
We evaluate the fidelity of individual object generation from single-view inputs. As visualized in~\autoref{demo_object}, the baseline MMGDreamer tends to produce over-smoothed and distorted geometries, failing to capture complex topological structures. For instance, the generated bookshelf appears to be melted with irregular compartments, and the desk lacks distinct boundaries for its drawers.
In contrast, FlowScene reconstructs precise geometries characterized by sharp edges and fine-grained details, alongside high-fidelity textures, faithfully preserving the structural integrity of the input images. As evidenced by examples of the wardrobe and bed, our method accurately renders intricate visual attributes, such as wood grain patterns and fabric folds. Consequently, FlowScene generates 3D objects that surpass the baselines in both geometric precision and appearance quality.

\section{Implementation Details}
\label{imp_detail}
\paragraph{Software and Hardware}
\begin{table}[!t]
\caption{\textbf{System and Software Configuration.}}
\centering
\scalebox{1}{
\begin{tabular}{lc} 
\toprule
\multicolumn{2}{c}{\textbf{System \& Hardware Overview}}  \\ 
\midrule
\multirow{1}{*}{CPU} & AMD EPYC 7702 64-Core             \\
GPU                  & 8 $\times$NVIDIA A100 GPU    \\
Memory               & 1T RAM                                 \\
Operating System     & Ubuntu 22.04 LTS                     \\
CUDA Version         & 12.8                                   \\
NVIDIA Driver        & 570.133.20                             \\
ML Framework         & Python 3.10.0  Pytorch 2.1.0          \\

\midrule
\multicolumn{2}{c}{\textbf{GPU Specifications}}  \\ 
\midrule
CUDA Cores& 6912 \\
Memory Capacity& 80GB \\
\bottomrule
\end{tabular}}
\label{tab: hardware}
\end{table}
All experiments are conducted on a computing system configured with an AMD EPYC 7702 64-core processor and 8×NVIDIA A100 GPUs, as detailed in~\autoref{tab: hardware}. Note that FlowScene is able to be trained with only one GPU. The system runs Ubuntu 22.04 LTS with CUDA 12.8, NVIDIA Driver 570.133.20, and uses Python 3.10.0 with PyTorch 2.1.0 as the machine learning framework. For rendering 3D indoor scenes, we employ Blender 4.1 with the CYCLES engine. The rendering settings include a Noise Threshold of 0.001, a max samples step of 300, and RGBA color mode with a 16 color depth. All rendering parameters are kept consistent across qualitative comparisons.


\paragraph{Dataset Details}
We conduct experiments on the publicly available SG-FRONT dataset~\cite{zhai2024commonscenes}, extended scene graph annotations of the 3D-FRONT dataset~\cite{fu20213d}. SG-FRONT encompasses 15 relationship types across three categories: spatial (e.g., \texttt{left/right of}, \texttt{bigger/smaller than}, \texttt{taller/shorter than}), support (e.g., \texttt{above}, \texttt{standing on}), and style (e.g., \texttt{the same material as}, \texttt{the same style as}). These diverse relationships allow for modeling spatial and semantic interactions among objects. 
In total, there are 4041 bedroom scenes, 900 dining room scenes, and 813 living room scenes. 
We follow the data split used in MMGDreamer~\cite{yang2025mmgdreamer}.
Unless otherwise specified, all ablation studies are evaluated on the full test set aggregated across all room types, rather than reported separately for each room category.
For the training-free language-based methods, we construct input instructions from scene graphs based on Instructscene~\cite{lin2024instructscene}.

\paragraph{Model Architectures}
All three branches adopt Transformer-based flow matching backbones similar to Trellis~\cite{xiang2024structured}, and incorporate a triplet-GCN~\cite{johnson2018image} as the core of the \textsc{InfoExchangeUnit} to enable iterative message passing. As illustrated in~\autoref{fig:gcn}, each branch instantiates a modality-specific unit variant with a dedicated projector.
All \textsc{InfoExchangeUnit}s use a 5-layer triplet-GCN backbone ($L=5$) with MLP hidden dimension $d_{\text{hidden}}=256$. Graph convolutional parameters are not shared across branches. In all branches, modality-specific denoising states are first projected to a unified dimension $d_{\text{gcn}}=64$, and then concatenated with graph node features before being processed by the triplet-GCN.
The denoisers of layout and shape branches $\Theta_{(\mathcal{B},\mathcal{S})}$ share a lightweight backbone configuration with 16 Transformer blocks, a hidden dimension of 512, and 16 attention heads. The layout branch employs the \textsc{LayoutExchangeUnit}, where bounding box attributes are projected to $d_{\text{gcn}}=64$ via a linear layer, and then concatenated with node features and processed by a triplet-GCN to produce node-wise layout conditioning.
The shape branch operates on a $16^3$ voxel grid and employs the \textsc{ShapeExchangeUnit}. Its projector consists of two 3D convolution layers with channels $32 \rightarrow 64$, followed by downsampling with stride 2 and stride 4, flattening, and a linear projection to $d_{\text{gcn}}=64$. The projected shape embeddings are concatenated with graph node features to generate shape conditioning.
The texture branch $\Theta_{\mathcal{T}}$ captures high-frequency details at $64^3$ resolution using a Sparse Transformer backbone wrapped with Sparse ResNet blocks, and instantiates the \textsc{TextureExchangeUnit}. Its sparse projector uses sparse 3D convolutions with channels $32 \rightarrow 64$, followed by sparse downsampling with factors 2 and 4, sparse-to-dense flattening, and a linear projection to $d_{\text{gcn}}=64$. The projected texture embeddings are concatenated with graph node features for graph-based texture conditioning.

\paragraph{VQ-VAE Details for Shape and Texture Branches}
Following Trellis~\cite{xiang2024structured}, both the shape and texture branches employ VQ-VAE models to encode high-dimensional structured representations into compact discrete latents.
\textbf{Shape VQ-VAE ($\Phi$).}
The shape branch adopts a 3D convolutional VQ-VAE to encode sparse voxelized geometry into discrete latent codes, as illustrated in~\autoref{fig:process}~\textcolor{cvprblue}{B}. The encoder $\Phi_E$ consists of a hierarchy of 3D convolutional and residual blocks with channel sizes ${32, 128, 512}$, mapping input voxel grids to a latent embedding space of dimension $8$. We use a codebook of size $1024$ for vector quantization. We adopt the standard VQ objective with an exponential moving average (EMA) codebook update and set the commitment loss weight to $0.25$. The decoder $\Phi_D$ mirrors the encoder with reversed channel ordering ${512, 128, 32}$ to reconstruct voxelized geometry from the discrete latent codes.
\textbf{Texture VQ-VAE ($\Psi$).}
The texture branch employs a structured latent VQ-VAE to encode multi-view appearance features anchored to object geometry, as illustrated in~\autoref{fig:process}~\textcolor{cvprblue}{C}. Following Trellis, multi-view image features are first extracted using a frozen DINOv2 encoder and aggregated into voxel-aligned feature volumes. The encoder $\Psi_E$ encodes these structured feature volumes into discrete latents of dimension $8$ with a codebook size of $1024$, using the same EMA-based vector quantization scheme and commitment weight of $0.25$. The decoder $\Psi_D$ is implemented as a high-capacity transformer-based mesh and appearance decoder, parameterized with 12 attention blocks, 12 attention heads, and a model dimension of 768, to reconstruct high-resolution texture representations conditioned on the discrete texture latents.

\paragraph{Training Setup and Inference}
We train the three branches of FlowScene independently using separate denoisers and exchange units, each optimized with the objective $\mathcal{L}_{\mathrm{GRF}}$. Each branch is optimized for 82k steps with the AdamW optimizer. A batch size of 196 is applied uniformly across all three branches. We employ a piecewise constant learning rate schedule, initialized at $1 \times 10^{-4}$ and decayed to $5 \times 10^{-5}$ and $1 \times 10^{-5}$ at 35k and 70k steps, respectively.To encode multimodal node information, we utilize a pre-trained DINOv2 encoder~\cite{oquab2023dinov2} for visual features and a frozen CLIP ViT-B/32 encoder~\cite{radford2021learning} for textual descriptions. Additionally, to foster robustness against diverse user inputs, we randomly mask 20\% of the textual and visual modalities within the graph nodes during both training and inference. For sampling, we adopt an Euler-based ODE solver enhanced with interval Classifier-Free Guidance (CFG), following Trellis~\cite{xiang2024structured}. Each branch performs inference using Algorithm~\ref{alg:grf_infer}. We utilize $K=25$ inference steps with a CFG strength of 5.0.

\begin{figure}
    \centering
    \includegraphics[width=1\linewidth]{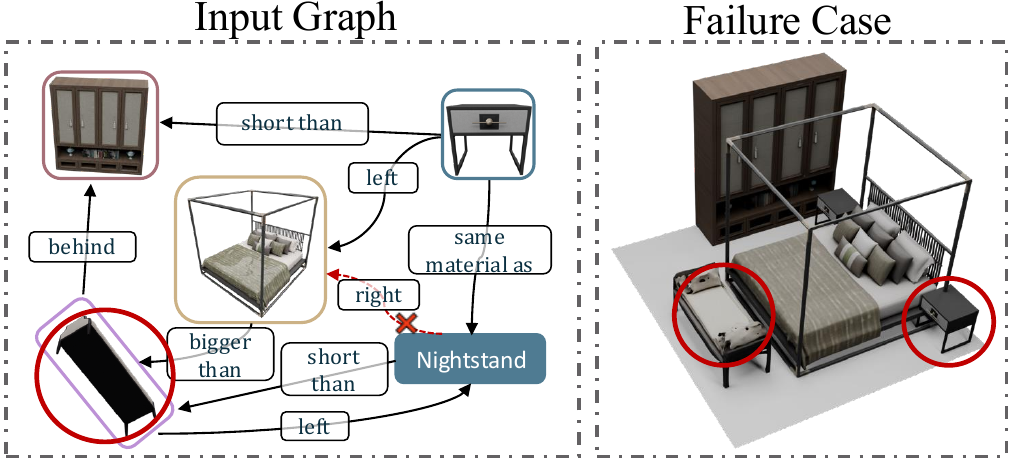}
    \caption{\textbf{Failure case.} The left panel shows the input multimodal scene graph, while the right panel shows the generated failure case. Red cross marks indicate removed relationships.} 
    \label{fig:sup_failure_case}
\end{figure}

\section{FPVScore Criteria Details}
\label{fpv_detail}
We implement FPVScore~\cite{huang2025video} using the GPT-4o model with the prompt template illustrated in~\autoref{fig:prompt-gpt}, performing a joint comparative evaluation of five distinct methods based on a shared text description. We present the LLM with 12 static multi-view images rendered from evenly spaced viewpoints around the scene, providing a 360-degree visual coverage. Each method is evaluated using the same set of viewpoints to ensure fair comparison.
We extend the standard criteria of Prompt Adherence (PA) and Layout Correctness (LC) by introducing Visual Quality (VQ) to assess visual fidelity and texture realism, and Style Consistency (SC) to evaluate the coherence of materials and design themes, specifically enforcing material uniformity across identical object instances. Finally, the model analyzes these visual inputs and assigns a rank from 1 (worst) to 5 (best) for each method across all five dimensions, including Overall Preference (OP).

\section{Perceptual Study Details}
\label{userstudy}
\begin{figure*}%
    \centering
    \includegraphics[width=0.95\linewidth]{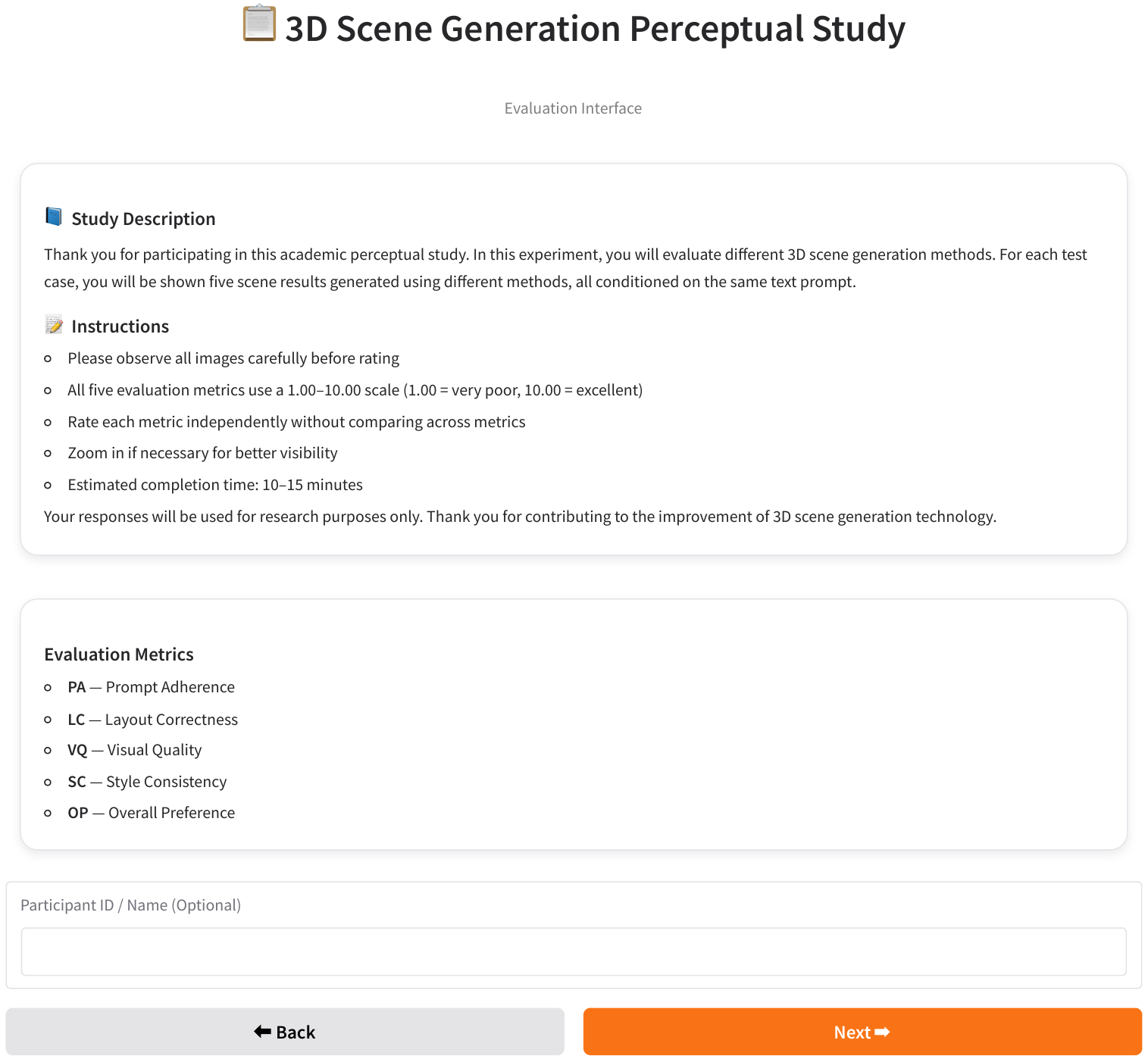}
    \caption{
    \textbf{Perceptual Study Interface: Instructions and Metrics.} This introductory page is presented to participants at the start of the perceptual study. It outlines the study objectives, provides detailed instructions for viewing and rating, and clearly defines the five evaluation metrics~(PA, LC, VQ, SC, OP).
    }
    \label{fig:per_study_inter_first}
\end{figure*}

\begin{figure*}%
    \centering
    \includegraphics[width=1\linewidth]{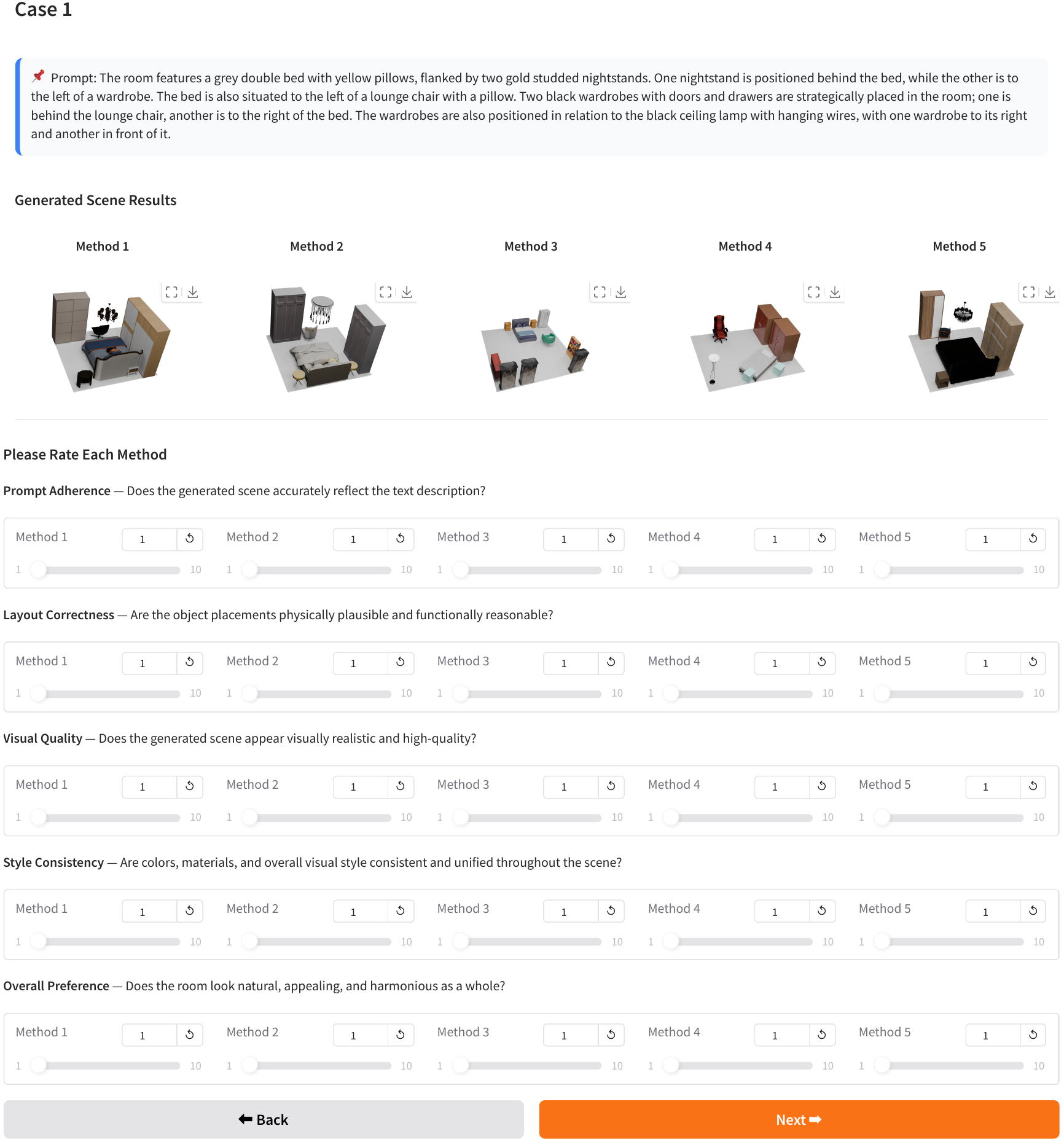}
    \caption{
    \textbf{Perceptual Study Interface: Example Page.} Participants are presented with a detailed text prompt (top) and the corresponding generated 3D scenes from five different methods. Users interactively rate each method using sliding scales (1.00-10.00) across the five defined criteria, allowing for a direct and comprehensive comparison of generation quality.
    }
    \label{fig:per_study_inter_example}
\end{figure*}

We conducted a perceptual study involving 25 voluntary participants across 20 distinct scenes using a custom web-based interface. Upon entry, participants were briefed on the detailed instructions and evaluation criteria via the landing page, as shown in \autoref{fig:per_study_inter_first}. Subsequently, they navigated to the evaluation page in \autoref{fig:per_study_inter_example} to assess the results. For each scene, outputs from five different methods were presented in a randomized order to prevent bias. Participants rated each method independently using a scale from 1.00 to 10.00 across five key metrics.

\section{Application Interface}
\label{interface}

\begin{figure*}[h]%
    \centering
    \includegraphics[width=0.82\linewidth]{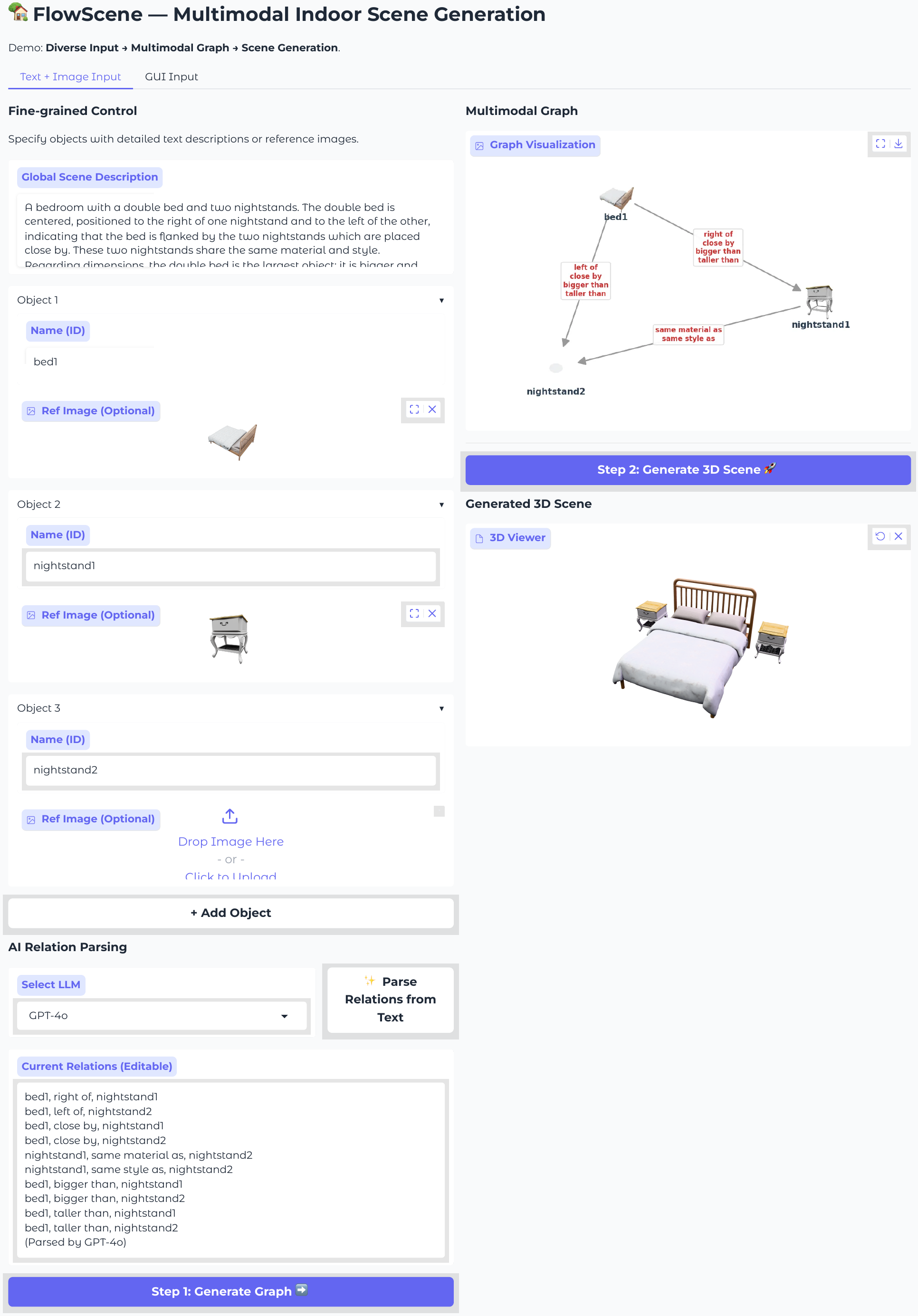}
    \caption{
    \textbf{The Interface of Language Description with Images.}
    }
    \label{fig:demo_llm}
\end{figure*}

\begin{figure*}[h]%
    \centering
    \includegraphics[width=0.85\linewidth]{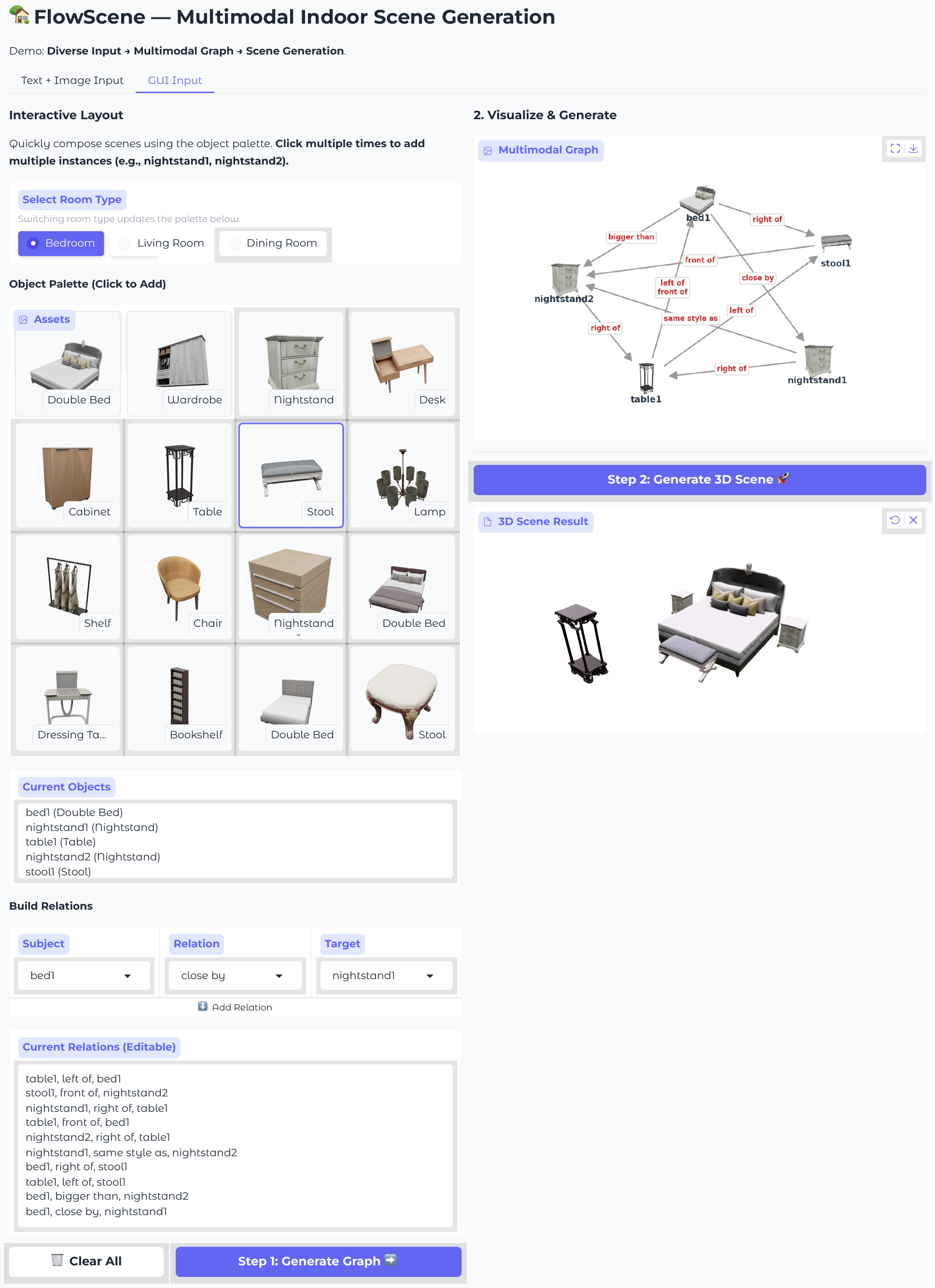}
    \caption{
    \textbf{The Interface of GUI.}
    }
    \label{fig:demo_gui}
\end{figure*}

We demonstrate the whole workflow with two application modes. \textbf{(i)} In a language-driven mode, users provide natural-language descriptions of objects and their spatial and semantic relations, optionally augmented with reference object images as visual appearances. As shown in~\autoref{fig:demo_llm}, an LLM such as DeepSeek~\cite{guo2025deepseek} or GPT-4o~\cite{achiam2023gpt} can be employed to parse the text into a scene graph, which is then converted into a multimodal graph by attaching the uploaded images to the corresponding nodes. FlowScene runs as a backend on this graph and generates a high-fidelity textured 3D scene that follows the described configuration. \textbf{(ii)} In an interactive GUI mode, also illustrated in~\autoref{fig:demo_gui}, users first choose a room type, after which the interface presents object candidates with images from multiple viewpoints. Users can select objects by clicking on these images or by specifying textual descriptions when they do not wish to fix the visual appearance. They can then define optional relations between object pairs. The GUI composes these choices into a multimodal graph, and again, FlowScene finally generates a 3D scene consistent with this graph. 
\emph{We provide two videos named \texttt{language-to-scene.mov} and \texttt{gui-to-scene.mov} in the Supplementary Material zip files.}
From a long-horizon perspective, FlowScene with scalable training on more data could generate textured objects from arbitrary given object images, allowing users to upload their desired images.

\begin{figure*}[!t]
    \centering
    \begin{center}
    \begin{tcolorbox} [top=2pt,bottom=2pt, width=\linewidth, boxrule=1pt]
    {\footnotesize {\fontfamily{zi4}\selectfont
    \textbf{FPVScore Prompt:}

Task: Compare the room layout rationality and overall performance of five methods, all generated from the same text description. 
From top to bottom, each row shows 12 static views sampled around the generated scene, providing a 360-degree multi-view coverage for each method.
Decide which method performs best according to the criteria below.

Text Description: \{text\_description\}

\vspace{2mm}

Instructions:

1. Prompt Adherence (PA) \\
Does the generated scene accurately reflect the text description? \\
Check whether all described objects are present and correctly represented.

2. Layout Correctness (LC)\\
Is the room design physically plausible and functional? \\
Evaluate if the layout supports practical use, space efficiency, and proper object functionality. Consider object positions, orientations, and user convenience.

3. Visual Quality (VQ) \\
Assess the visual realism and rendering quality of the generated scene. \\
Consider texture fidelity, lighting realism, material smoothness, and absence of artifacts or distortions.

4. Style Consistency (SC) \\
Evaluate whether the overall aesthetic style is coherent across the scene. \\
Check if color palettes, material usage, and design patterns are consistent and match the intended scene theme.

5. Overall Preference (OP) \\
Does the room layout look natural, appealing, and harmonious as a whole? \\
Consider the holistic impression combining semantics, layout, visuals, and style.

\vspace{2mm}

Evaluation process: \\
Carefully examine the multi-view images of all five 3D scenes. Focus on one criterion at a time and make independent judgments for each.

\vspace{2mm}

Output format: \\
Provide a clear, concise analysis for each criterion. \\
After the analyses, assign ranks (1$–$5) to each method per criterion (1 = worst, 5 = best).

Summarize your final ranking in the format: 
\textlangle rank for criterion 1 \textrangle \textlangle rank for criterion 2 \textrangle \textlangle rank for criterion 3 \textrangle \textlangle rank for criterion 4 \textrangle \textlangle rank for criterion 5 \textrangle\ for each method.

\vspace{2mm}
\medskip
Example:

\textit{Analysis:} \\
1. Prompt Adherence: The first one ...; The second one ...; The third one ...; The fourth one ...; The fifth one ... \\
2. Layout Correctness: The first one ...; The second one ...; The third one ...; The fourth one ...; The fifth one ...\\
3. Visual Quality: The first one ...; The second one ...; The third one ...; The fourth one ...; The fifth one ... \\
4. Style Consistency: The first one ...; The second one ...; The third one ...; The fourth one ...; The fifth one ... \\
5. Overall Preference: The first one ...; The second one ...; The third one ...; The fourth one ...; The fifth one ...

\medskip
\textit{Final answer:} \\
The first one: x x x x x \\
The second one: x x x x x \\
The third one: x x x x x \\
The fourth one: x x x x x \\
The fifth one: x x x x x \\

}
\par}
    \end{tcolorbox}

    \end{center}
    \caption{\textbf{Prompt template} for evaluating FPVScore~\cite{huang2025video} with GPT-4o~\cite{achiam2023gpt}.}
    \label{fig:prompt-gpt}
\end{figure*}


\section{Limitations}
\label{limitation}
Our current implementation focuses primarily on synthetic indoor environments (e.g., 3D-FRONT~\cite{fu20213d}) and has not yet been evaluated on outdoor scenes. Additionally, the generation quality is intrinsically dependent on the precision of the input multimodal graph; consequently, inaccuracies or ambiguity from upstream graph constructors (LLMs/VLMs) may propagate to the final scene structure. 
For example, when the input image is captured from a limited or biased viewpoint, the model may generate imperfect object shapes. 
Moreover, missing key spatial relationship edges in the graph may reduce layout constraints, which can result in object interpenetration.
We present such failure cases in~\autoref{fig:sup_failure_case}.
Future work will aim to extend the model’s applicability to broader domains and enhance its robustness against noisy graph inputs.

\end{document}